\documentclass[10pt,twocolumn,letterpaper]{article}

\usepackage[pagenumbers]{cvpr} 

\usepackage[dvipsnames]{xcolor}

\newcommand{\algname}{$\texttt{Forward-INF}$~}
\usepackage{caption}
\usepackage{algorithm}
\usepackage{algpseudocode}
\usepackage{multirow}

\definecolor{cvprblue}{rgb}{0.21,0.49,0.74}
\usepackage[pagebackref,breaklinks,colorlinks,citecolor=cvprblue]{hyperref}

\def\paperID{16375} 
\def\confName{CVPR}
\def\confYear{2024}

\title{The Mirrored Influence Hypothesis: Efficient Data Influence Estimation by Harnessing Forward Passes}

\author{Myeongseob Ko$^1$\hspace{1.5em}
Feiyang Kang$^1$\hspace{1.5em}
Weiyan Shi$^2$\hspace{1.5em}
Ming Jin$^1$\hspace{1.5em}
Zhou Yu$^2$\hspace{1.5em}
Ruoxi Jia$^1$\\[1ex] 
$^1$Virginia Tech\hspace{1.5em} $^2$Columbia University\\
{\tt\small \{myeongseob, fyk, jinming, ruoxijia\}@vt.edu, \tt\small \{ws2634, zy2461\}@columbia.edu}
}

\begin{document}
\maketitle
\begin{abstract}
Large-scale black-box models have become ubiquitous across numerous applications. Understanding the influence of individual training data sources on predictions made by these models is crucial for improving their trustworthiness. Current influence estimation techniques involve computing gradients for every training point or repeated training on different subsets. These approaches face obvious computational challenges when scaled up to large datasets and models.

In this paper, we introduce and explore the Mirrored Influence Hypothesis, highlighting a reciprocal nature of influence between training and test data. Specifically, it suggests that evaluating the influence of training data on test predictions can be reformulated as an equivalent, yet inverse problem: assessing how the predictions for training samples would be altered if the model were trained on specific test samples. Through both empirical and theoretical validations, we demonstrate the wide applicability of our hypothesis. Inspired by this, we introduce a new method for estimating the influence of training data, which requires calculating gradients for specific test samples, paired with a forward pass for each training point. This approach can capitalize on the common asymmetry in scenarios where the number of test samples under concurrent examination is much smaller than the scale of the training dataset, thus gaining a significant improvement in efficiency compared to existing approaches.
We demonstrate the applicability of our method across a range of scenarios, including data attribution in diffusion models, data leakage detection, analysis of memorization, mislabeled data detection, and tracing behavior in language models. 
Our code is available at https://github.com/ruoxi-jia-group/Forward-INF.
\end{abstract}    

\section{Introduction}
\label{sec:intro}

As the popularity of large-scale, black-box machine learning models continues to surge across diverse applications, the need for transparency---an understanding of the factors driving their predictive behaviors---becomes increasingly critical. These models are learned from training data, and as such, an important step towards achieving transparency lies in estimating the influence of individual training data points on the model's predictions.

Extensive research on training data influence estimation has been conducted over the years~\citep{koh2017understanding,pruthi2020estimating,park2023trak,ilyas2022datamodels,jia2019towards}. Despite the diversity of techniques, they all fundamentally revolve around a central idea of assessing the counterfactual impact of a training data source:

\begin{quote}
   \emph{How would the prediction on specific test points change if we removed a training source?} (\textbf{P1})
\end{quote}

One line of approaches~\citep{ilyas2022datamodels} focuses on the direct evaluation of the counterfactual impact, i.e., by retraining a model on the set excluding the training source and measuring the change in the prediction. Besides the obvious computational overhead, such evaluation results suffer from a low signal-to-noise ratio due to the stochasticity in widely-used learning algorithms and are largely inconsistent across different runs~\citep{wang2023data}. To magnify the change caused by removing a single source, existing techniques mostly involve retraining models on smaller subsets of the training data and measuring a source's influence by aggregating its contribution to different subsets not containing the source~\citep{feldman2020neural,jia2019efficient,ghorbani2019data}. While these methods produce more consistent influence scores, they are infeasible for large-scale models.

\begin{figure}[t!]
\centering
\includegraphics[width=0.95\linewidth]{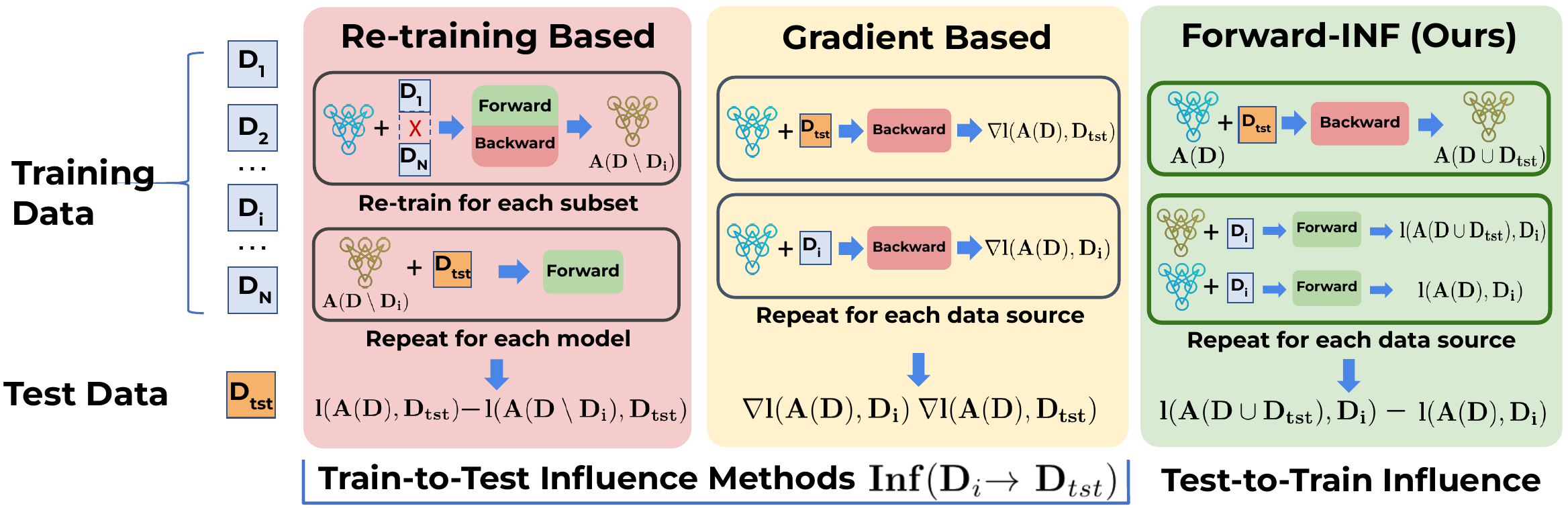}
\caption{Overview of our approach and comparison with prior work which can be generally categorized into re-training-based methods and gradient-based methods. The former requires re-training models on many different subsets of training data~\citep{ilyas2022datamodels,jia2019efficient,ghorbani2019data,feldman2020neural}. The latter calculates the influences based on training data gradients (\texttt{TracIn}~\citep{pruthi2020estimating} is illustrated as an example in the second column). Our proposed method \algname features only forward pass computation for each training point, offering significant efficiency improvement.}
\label{fig:overview}
\vspace{-1em}
\end{figure}

Another line of approaches bypasses the need for re-training by estimating the influence through the final trained model or intermediate checkpoints reached during the training process~\citep{koh2017understanding,pruthi2020estimating}. In particular, they evaluate a training point's influence using the corresponding gradient of the final model or checkpoints, which effectively represents the local changes made by introducing the point into training. However, calculating gradients is not only time-intensive but also memory-inefficient compared to forward pass~\citep{malladi2023fine}. In our tests, it took up to 10.92 times longer and used 5.36 times more memory than inference for the Vision Transformer \texttt{vit-b-32}~\citep{dosovitskiy2020image}. Furthermore, identifying the most influential training point requires computing the gradient for \emph{every single} training data point. Combined, these challenges hinder the efficient determination of data influence, especially for large-scale datasets and large models. Figure~\ref{fig:overview} highlights the conceptual difference between existing approaches and we will defer the detailed discussion of Related Work to Appendix~\ref{sec:Related}. 

Given the discrepancy in efficiency between gradient calculation and inference, an intriguing yet uncharted question arises: \emph{Can we maximize the usage of forward pass when estimating influence?} We will introduce a technique that exploits the considerable efficiency gap between forward and backward passes, especially when applied to all training points. This method is anchored on the following Mirrored Influence Hypothesis.


\noindent
\textbf{The Mirrored Influence Hypothesis.} \emph{The train-to-test influence characterized by the problem (P1) is correlated with the test-to-train influence characterized by (P2) in which the role of training and test points is swapped:} 

\begin{quote}
     \emph{How would the prediction on a training source change if the model was trained on specific test points?} (\textbf{P2})
\end{quote}

\begin{figure}[t!]
\centering
\includegraphics[width=0.8\linewidth]{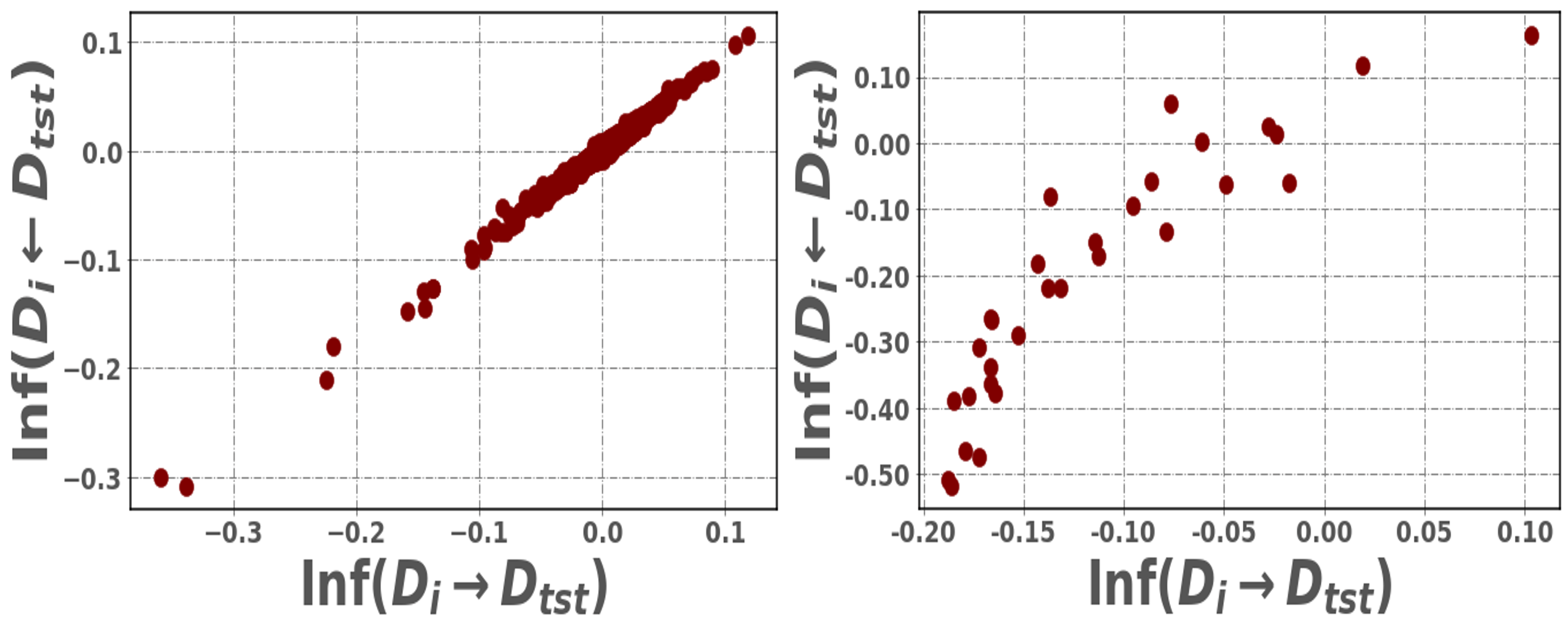}
\caption{We observe high correlation between train-to-test influence $\text{Inf}(D_i\rightarrow D_\text{tst})$ and test-to-train influence $\text{Inf}(D_i\leftarrow D_\text{tst})$. The average Pearson Correlation is 0.9673 for logistic regression and 0.8851 for CNN trained on CIFAR-10.
} 
\vspace{-1em}
\label{fig:mirrored-inf}
\end{figure}

More formally, consider a training dataset comprising $N$ data sources, $D_\text{trn}=D_1\cup \ldots \cup D_N$ and a test set $D_\text{tst}$. Let $\mathcal{A}$ denote the learning algorithm which takes a dataset as input and returns a model, and let $\mathcal{L}$ be a loss function. The \emph{train-to-test} influence characterized by \textbf{P1} can be expressed as
\begin{equation}
\resizebox{0.9\columnwidth}{!}{
$\begin{aligned}
\label{eqn:train-to-test}
    \text{Inf}(D_i\rightarrow D_\text{tst}) = \mathcal{L}(\mathcal{A}(D_\text{trn}),D_\text{tst}) - \mathcal{L}(\mathcal{A}(D_\text{trn}\setminus D_i),D_\text{tst}).
\end{aligned}$
}
\end{equation}
On the other hand, the \emph{test-to-train} influence characterized by \textbf{P2} can be written as
\begin{equation}
\resizebox{0.9\columnwidth}{!}{
$\begin{aligned}
\label{eqn:test-to-train}
    \text{Inf}(D_i\leftarrow D_\text{tst}) = \mathcal{L}(\mathcal{A}(D_\text{trn}\cup D_\text{tst}),D_i) - \mathcal{L}(\mathcal{A}(D_\text{trn}),D_i).
\end{aligned}$
}
\end{equation}
We find that $\text{Inf}(D_i\rightarrow D_\text{tst})$ is \emph{highly correlated} 
with $\text{Inf}(D_i\leftarrow D_\text{tst})$ for $i\in\{1,\ldots,N\}$. Figure~\ref{fig:mirrored-inf} illustrates the correlation for convex and non-convex models trained on the CIFAR-10 dataset and we defer results on other datasets and models to the Appendix~\ref{app: more empirical results} due to the similarity in their trends.

\noindent
\textbf{Leveraging the Hypothesis.} In most data influence estimation applications, the size of the test set is typically much smaller than the training set (\(|D_{\text{tst}}|\) $<<$\(|D_{\text{trn}}|\)). For example, when identifying influential training points for a specific model prediction, \(D_{\text{tst}}\) represents just a single test point. Similarly, in detecting low-quality training data, \(D_{\text{tst}}\) is a small, clean reference set, usually less than 1\% of the entire training set~\citep{zeng2021adversarial,xiang2021post,just2023lava}. The Mirrored Influence Hypothesis enables leveraging this size asymmetry between training data and test samples under concurrent examination to develop more efficient influence estimation algorithms. This hypothesis allows for a shift in approach: from calculating the train-to-test influence of training data, which requires locally updating the model for each training point, to assessing the test-to-train influence. In practice, this means updating the model for the relatively few test samples and conducting forward passes on training samples. This shift effectively applies the more computationally intensive process (the backward pass) to the smaller scale test set, while the computationally lighter task (the forward pass) is applied to the larger training data, optimizing overall efficiency.

 \noindent
\textbf{Evaluation.} Leveraging the insights articulated, we develop a new influence estimation algorithm. We evaluate its performance across diverse applications handling image data, such as detecting mislabeled data, identifying data leakage, analyzing memorization, and attributing data in diffusion models. To further demonstrate its wide-ranging applicability, we extended its use to tracing behavior in language models. This new method not only showcases promising utility but also offers significantly faster performance compared to traditional train-to-test influence focused techniques. For instance, our method detects data leakage with 100\% accuracy on CIFAR-10, surpassing the Influence Function~\citep{koh2017understanding} (30\% detection rate) and TracIn~\citep{pruthi2020estimating} (0\% detection rate) by being 30 and 40 times faster, respectively. 





\section{Delve Into the Hypothesis}
\label{sec:Approach}

In this section, we empirically assess the Mirrored Influence Hypothesis as applied to different datasets and models. Then, we study different influence approximators: Influence Function~\citep{koh2017understanding} and TracIn~\citep{pruthi2020estimating}, and show that our hypothesized correlation between forward and backward influences holds naturally under the assumptions made by these approximators. 




\subsection{Empirical Study}
\label{sec: empirical study}

We investigate the Hypothesis in both deterministic and stochastic learning settings, due to the distinct noise levels inherent in the influence scores of these two scenarios. Specifically, both train-to-test and test-to-train influences depend on the chosen learning algorithm \(\mathcal{A}\). Algorithms like stochastic gradient descent (SGD) introduce randomness, such as through random mini-batch selection, which affects the training process and, consequently, the trained models. This randomness can cause variations in the losses evaluated on these models, ultimately impacting the influence scores, which are derived from these losses, as seen in Eqn. (\ref{eqn:train-to-test}) and (\ref{eqn:test-to-train}). 

Particularly, the influences for individual training and test points, which are determined by the exclusion of a training point or the addition of a test data point into the training set, are typically small and can be heavily impacted by the stochasticity of the learning process. For example, prior research~\citep{wang2022data} shows that the variability in an individual point's influence due to learning stochasticity often largely surpasses their magnitude. This means that even when evaluating the same type of influence, there can be a significant inconsistency in the rankings of different data points' influence scores across different algorithm runs. Hence, to meaningfully analyze the correlation between train-to-test and test-to-train influences, it is critical to focus on settings where the signal strength in both types of influence scores is substantial enough to withstand the overshadowing effect of noise.

\noindent\textbf{A Near-Noiseless Setting.} We start with a near-noiseless setting where the model being trained is strongly convex and the learning algorithm is deterministic. With long enough iterations, the final model \(\mathcal{A}(D)\) is guaranteed to converge to the vicinity of the global minima for any training dataset \(D\). Specifically, we train a logistic regression model using L-BFGS~\citep{liu1989limited} with L2 regularization. Due to the low noise in this setting, we can investigate the \textbf{point-to-point influence}, i.e., $|D_i|=|D_\text{tst}|=1$.
We use Pearson correlation and Spearman rank-order correlation coefficient to measure the correlation between the two influences. As we are mainly interested in a test point that has a high loss (e.g., in the application of debugging a misclassified test point), we select 10 test points that have the highest loss and take an average over their correlation scores.

\noindent\textbf{A Noisy Setting.} As discussed, training non-convex models using stochastic learning algorithms inherently produces a high level of noise in the influence between individual training and test points. Specifically, we train a convolutional neural network (CNN) using SGD with a learning rate of 0.01. To mitigate the noise, our initial approach is to average the influence scores over multiple runs of the learning algorithm. However, the number of runs required to sufficiently reduce the noise is prohibitively demanding. To effectively analyze the correlation between two types of influences, we design experiments where the influence scores across different $D_i$ are more significant in magnitude, thus reducing the chance of these scores being overwhelmed by noise. Specifically, we assess the \textbf{group-to-group influence}, i.e., the impact of various groups of training points on a test set. Effectively, modifying a group of points in the training set would result in more substantial changes in loss, thereby making the influence scores more indicative.  
To amplify the effect of each group's removal or addition, we randomly mislabel 50\% of samples into different classes and assign varying mislabeling ratios to each group. We use 30 different training groups with mislabeling ratios ranging linearly from 0\% to 100\%. $D_\text{tst}$ consists solely of clean samples.

\noindent\textbf{Result.} In Table~\ref{tab:corr}, we show Pearson and Spearman rank-order correlations between $\text{Inf}(D_i\rightarrow D_\text{tst})$ and $\text{Inf}(D_i\leftarrow D_\text{tst})$ for both settings.
For the noiseless setting, we observe high correlation scores (i.e., greater than 0.96 in terms of Pearson Correlation) across different datasets. On the other hand, the Spearman correlation is relatively lower. As illustrated by Figure~\ref{fig:mirrored-inf}, there exists a high-density region with many similar values, which could lead to a lot of tied ranks. These ties can disrupt the monotonic relationship that Spearman correlation seeks to measure.

In the noisy setting with group-to-group influence, the correlation between the two types of influences remains high. Notably, while the Pearson correlation exhibits a decrease in comparison to the noiseless scenario, Spearman correlation retains a high value. This discrepancy arises partly because Pearson correlation, which relies on actual data values, is more susceptible to noise. In contrast, Spearman correlation employs ranks rather than raw values, which inherently provides resistance to the distorting effects of noise. Moreover, as depicted in Figure~\ref{fig:mirrored-inf}, the relationship between train-to-test and test-to-train influences maintains its correlation, albeit with a diminished linear characteristic. The pronounced linear correlation for point-to-point influences in the noiseless setting aligns with the theoretical insights discussed in the subsequent subsection. These insights indicate that under minor perturbations to the training set, such as the alteration of a single data point, it can be shown that training and test data have symmetrical roles and the two types of influences are empirically equivalent. Conversely, in the context of the group-to-group influence, the removal or addition of entire data groups leads to more significant model perturbations, which are not adequately described by current theoretical frameworks. An in-depth theoretical exploration of the correlations between the two influences under general conditions is outside this paper's scope and presents an intriguing avenue for future research.

\begin{table}[t!]
\centering
\resizebox{0.9\linewidth}{!}{
\begin{tabular}{llccc}
\textbf{Setting}     & \textbf{Metric} & \textbf{MNIST} & \textbf{FMNIST} & \textbf{CIFAR-10} \\ \cmidrule[0.4pt](l{0.525em}r{0.525em}){1-1} \cmidrule[0.4pt](l{0.525em}r{0.525em}){2-2} \cmidrule[0.4pt](l{0.525em}r{0.525em}){3-3} \cmidrule[0.4pt](l{0.525em}r{0.525em}){4-4} \cmidrule[0.4pt](l{0.525em}r{0.525em}){5-5}
\multirow{2}{3cm}{Near-Noiseless \\(point-to-point inf)}     & Pearson         & 0.9975          & 0.9939           & 0.9673           \\
                             & Spearman        & 0.9027          & 0.7274           & 0.8069           \\ \cmidrule[0.4pt](l{0.525em}r{0.525em}){1-1} \cmidrule[0.4pt](l{0.525em}r{0.525em}){2-2} \cmidrule[0.4pt](l{0.525em}r{0.525em}){3-3} \cmidrule[0.4pt](l{0.525em}r{0.525em}){4-4} \cmidrule[0.4pt](l{0.525em}r{0.525em}){5-5}
\multirow{2}{3cm}{Noisy\\(group-to-group inf)}  & Pearson         & 0.9640          & 0.9752           & 0.8551           \\
                             & Spearman        & 0.9907          & 0.9915           & 0.8848  \\
                             \cmidrule[0.9pt](l{0.525em}r{0.525em}){1-1} \cmidrule[0.9pt](l{0.525em}r{0.525em}){2-2} \cmidrule[0.9pt](l{0.525em}r{0.525em}){3-3} \cmidrule[0.9pt](l{0.525em}r{0.525em}){4-4} \cmidrule[0.9pt](l{0.525em}r{0.525em}){5-5}
\end{tabular}}
\caption{The evaluation of the Mirrored Influence Hypothesis in different settings with different datasets.}
\label{tab:corr}
\vspace{-1em}
\end{table}

\subsection{Validity of the Hypothesis for Influence Approximators}
In addition to the empirical study, we show that the hypothesized mirrored influence holds for well-known influence approximators: Influence functions~\citep{koh2017understanding} and TracIn~\citep{pruthi2020estimating}.

We begin by introducing the notations. For ease of exposition, we will examine the validity of the Hypothesis in the case where $|D_i|=1$ and $|D_\text{tst}|=1$. The argument naturally extends to more general cases. Specifically,
consider a training set consisting of $n$ samples 
$D_\text{trn} = \{z_1, z_2, ... z_n\}$, a given test sample $z_\text{tst}$,
and a neural network parameterized by $\theta \in \mathbb{R}^d$. 
Define the loss function $\mathcal{L}$ as the model's empirical risk on the training dataset $\mathcal{L}(\theta, D_\text{trn})=\frac{1}{n}\sum_{i=1}^n \ell(\theta, z_i)$, where $\ell(\theta, z_i)$ denotes the loss of a predictor parameterized by $\theta$ on the training sample $z_i$. The optimal model parameters are given by the following empirical risk minimization: $\hat{\theta}:=\arg\min_\theta \mathcal{L}(\theta, D_\text{trn})$.

\noindent\textbf{Influence Function (IF).} The idea of IF is to analyze the change in prediction loss when a training sample is up-weighted infinitesimally. In particular, if we perturb the weight of a sample $z$ from $1$ to $1+\varepsilon$, the new parameter on the perturbed training dataset can be given as $\hat{\theta}_{\varepsilon, z}=\arg\min_\theta \mathcal{L}(\theta, D)+\varepsilon \ell(\theta,z)$. 
With assumptions on the loss function being twice-differentiable and strictly convex, the influence of an infinitesimal perturbation of a training sample $z$ on the loss of a test sample $z_\text{tst}$ can be calculated as ~\citep{koh2017understanding}
\begin{align}\label{eqn:grad-delta}
   \left. \frac{d\ell(\hat\theta_{\varepsilon, z},z_\text{tst})}{d\epsilon}\right|_{\varepsilon=0} = -\nabla_\theta\ell(\hat\theta,z_\text{tst})^T H_{\hat\theta}^{-1}\nabla_\theta\ell(\hat\theta, z),
\end{align}
where $H_{\hat\theta}:=\nabla_\theta^2\ell(\hat\theta, D)$ denotes the Hessian. Since removing a point $z$ is equivalent to upweighting it by $\varepsilon=-\frac{1}{n}$, for $n$ sufficiently large and $\frac{1}{n}\rightarrow0$, one can approximate the train-to-test influence defined in Eqn. (\ref{eqn:train-to-test}) with its first-order Taylor approximation, which gives
\begin{equation*}
    \text{Inf}(D_i\rightarrow D_\text{tst})\approx \left(-\frac{1}{n}\right)\cdot \left[- \left. \frac{d\ell(\hat\theta_{\varepsilon, z},z_\text{tst})}{d\varepsilon}\right|_{\varepsilon=0}\right],
\end{equation*}
combining with Eqn. (\ref{eqn:grad-delta}), we have
\begin{align}\label{eqn:inf-rmov}
    \text{Inf}(D_i\rightarrow D_\text{tst}) \approx -\frac{1}{n}  \nabla_\theta\ell(\hat\theta,z_\text{tst})^T H_{\hat\theta}^{-1}\nabla_\theta\ell(\hat\theta, z)
\end{align}
Symmetrically, consider the \textit{alternative} of adding a test sample $z_\text{tst}$ with weight $\varepsilon$  to the training dataset. The model trained with the new objective will be $\hat\theta_{\varepsilon,z_\text{tst}}=\arg\min_\theta \mathcal{L}(\theta,D)+\varepsilon\ell(\theta,z_\text{tst})$. Similar to Eqn. (\ref{eqn:grad-delta}), the influence of training on the test sample $z_\text{tst}$ with an infinitesimal weight $\varepsilon$ on the prediction loss of a training sample $z$ can be calculated as 
\begin{align}\label{eq:inf-dual}
\left.\frac{d\ell(\hat\theta_{\varepsilon,z_\text{tst}}, z)}{d\varepsilon}\right|_{\varepsilon=0}  =-\nabla_\theta\ell(\hat{\theta}, z)^T H_{\hat{\theta}}^{-1}\nabla_\theta\ell(\hat\theta, z_\text{tst})
\end{align}
As adding a test point into the training set is the same as setting $\varepsilon =\frac{1}{n}$, following the same procedure, we can again linearly approximate the test-to-train influence in Eqn. (\ref{eqn:test-to-train}) by computing 
\begin{align}\label{eqn:inf-add}
    \text{Inf}(D_i\leftarrow D_\text{tst}) \approx -\frac{1}{n}\nabla_\theta\ell(\hat{\theta}, z)^T H_{\hat{\theta}}^{-1}\nabla_\theta\ell(\hat\theta, z_\text{tst})
\end{align}
Due to the fact that Hessian is symmetric by definition, Eqn. (\ref{eqn:inf-rmov}) and Eqn. (\ref{eqn:inf-add}) are equivalent. Then, 
we can observe that using influence functions to approximate \textit{influence} yields the same result for both train-to-test and test-to-train influences, which coincides with our Mirrored Influence Hypothesis. 

\noindent\textbf{TracIn.} TracIn~\citep{pruthi2020estimating} approximates the influence of a training sample $z^t_i$ on a testing sample $z_\text{tst}$ using a first-order approximation of the model and aggregating through multiple checkpoints during the training process:
\begin{equation}\label{eq:tracin}
    \text{Inf}(D_i\rightarrow D_\text{tst})\approx  \sum_{c\in \mathcal{C}} \eta_c\nabla_\theta\ell(\hat\theta_c, z_\text{tst})\cdot\nabla_\theta\ell(\hat\theta_c, z),
\end{equation}
where $c$ denotes an index of iteration during model training, $\mathcal{C}$ denotes the set of iteration indices where checkpoints are available, and $\eta_c$ and $\hat\theta_c$ represent the step size and the model weights at iteration $c$, respectively. Then, we also consider the alternative of estimating the influence on training sample $z$ by training the model on the test sample $z_\text{tst}$ at each checkpoint, which can be easily given as 

\begin{align}\label{eq:tracin'}
    \text{Inf}(D_i\leftarrow D_\text{tst})\approx \sum_{c\in \mathcal{C}} \eta_c\nabla_\theta\ell(\hat{\theta}_c, z)\cdot\nabla_\theta\ell(\hat{\theta}_c, z_\text{tst}) 
\end{align}
Again, with the TracIn approximator, the \textit{influence} between a training-test sample pair can be calculated from both directions and achieve the same result. Echoing our intuition, these results suggest the universal application of the proposed Hypothesis in influence approximators.


\section{\texttt{Forward-INF}: An Influence Approximation Algorithm Harnessing Forward Passes}
\label{sec:our approach}

In this section, we will present a new data influence estimation algorithm unlocked by the Hypothesis. This algorithm differs from existing methods by substituting the backward pass computations for individual training points with a forward pass, aiming to reduce computational costs.

Inspired by the Hypothesis, to rank the train-to-test influences among different training points, we can alternatively arrange them in order based on the test-to-train influences, as described in Eqn. (\ref{eqn:test-to-train}). Specifically, calculating $\text{Inf}(D_i\leftarrow D_\text{tst})$ involves first acquiring two models $\hat{\theta}=\mathcal{A}(D)$ and $\hat{\theta}_{+D_\text{tst}}=\mathcal{A}(D\cup D_\text{tst})$. We assume that $\hat{\theta}$ is available after training. As getting $\hat{\theta}_{+D_\text{tst}}$ by training on $D\cup D_\text{tst}$ from scratch can be expensive, we propose to use continual learning and obtain this model by updating the existing model $\hat{\theta}$ with $D_\text{tst}$. Denote the resulting model as $\tilde{\theta}_{+D_\text{tst}}$. In Appendix~\ref{app: more theoretical derivations}, we will show that when $n$ is large and $\hat{\theta}$ and $\hat{\theta}_{+D_\text{tst}}$ are close, continually updating $\hat{\theta}$ on $D_\text{tst}$ is a good approximation to training from scratch. In Appendix~\ref{app:analysis and details}, we will also empirically compare the resulting influence scores of the two settings. Finally, with the two models $\hat{\theta}$ and $\tilde{\theta}_{+D_\text{tst}}$, one can calculate the influence for each training source $D_i$ by two forward passes, which give $\mathcal{L}(\tilde{\theta}_{+D_\text{tst}},D_i)$ and $\mathcal{L}(\hat{\theta},D_i)$, and then take the difference.

\noindent\textbf{Implementation.} The pseudo-code is provided in Algorithm~\ref{alg:forwardinf}. We call this algorithm the \texttt{Forward-INF} algorithm because it implements forward passes on the training set. Note that this algorithm still applies backward passes on the test set. However, as typically, the training size is orders of magnitude larger than the number of test points being inspected concurrently, this algorithm is usually much faster than existing methods, like Influence Functions and TracIn, which apply backward passes on the training set (and the test set). 

Also, note that gradient ascent is implemented as default to update $\hat{\theta}$. This is because if the test sample is drawn from a similar distribution as the training data, the magnitude of the gradient $\left.\nabla_\theta\ell(\theta,z_\text{tst})\right|_{\theta=\hat\theta}$ tends to be small. Thus, employing gradient descent would introduce a small loss difference $\mathcal{L}(\hat{\theta}_K, D_i) - \mathcal{L}(\hat{\theta}, D_i)$. By contrast, gradient ascent is likely to produce a model underfitting for points similar to the test points, thus resulting in a larger loss difference for training points, which is beneficial for comparing their influences. However, it is observed in experiments that both gradient descent and ascent perform similarly well. A detailed ablation study on this aspect is deferred to Appendix~\ref{app:analysis and details}.

\begin{algorithm}
\caption{Forward-INF Algorithm}\label{alg:forwardinf}
\begin{algorithmic}[1]
\Require Training set $D_\text{trn}=D_1\cup\ldots\cup D_n$, Trained model $\hat{\theta}$, Target test samples $D_\text{tst}$, Number of continual learning iterations $K$, Learning rate $\alpha$
\Ensure \algname($D_i$) for $i=1,\ldots,n$
\State Initialize $\hat{\theta}_0 \leftarrow \hat{\theta}$
\For{$j = 1$ to $K$}
  \State Gradient ascent: $\hat{\theta}_{j+1} \leftarrow \hat{\theta}_j + \alpha \nabla_{\theta} \mathcal{L}(\hat{\theta}_j, D_\text{tst})$
\EndFor
\For{$D_i \subset D_\text{trn}$}
  \State Forward pass of $\hat{\theta}_K$ to get  $\mathcal{L}(\hat{\theta}_K, D_i)$
  \State Forward pass of $\hat{\theta}$ to get $\mathcal{L}(\hat{\theta}, D_i)$  
  \State \algname($D_i$) = $\mathcal{L}(\hat{\theta}_K, D_i) - \mathcal{L}(\hat{\theta}, D_i)$  
\EndFor
\end{algorithmic}
\end{algorithm}

\noindent\textbf{Hyperparameter.} The number of maximization iterations $K$ and the learning rate $\alpha$ can be tuned if one has access to or can create some ground-truth ``influential points.'' For example, one could use a subset of training data as test samples. Intuitively, the same training point would be most influential to the test. In this case, one can tune $K$ so that the duplicates in the training set (i.e., the ground-truth influential point) are assigned with the highest influence score.


\section{Application}
\label{sec:experiment}
In this section, we evaluate our approach to both vision and natural language processing (NLP) tasks. In particular, we apply our proposed method to the data influence estimation problem in diffusion models~\citep{wang2023data} (Section~\ref{sec: diffusion}), data leakage detection~\citep{barz2020we} (Section~\ref{sec: training data leakage subsection}),  analysis of memorization~\citep{feldman2020neural} (Section~\ref{sec: memorization}) as well as mislabeled data detection (Section~\ref{sec: mislabeled detection}).

We extend our method into the NLP task to showcase the performance in the context of a model behavior tracing task ~\citep{akyurek2022tracing} (Section~\ref{sec: model behavior tracing}). The scope of this study is to emphasize the method's \textbf{versatility across different applications} rather than to outperform existing \emph{application-specific} baselines in each case. However, we will compare our test-to-train influence calculation method with existing train-to-test influence methods, both characterized by their application-agnostic nature.

\subsection{Data Influence Estimation in Diffusion Model}
\label{sec: diffusion}

\begin{figure}[t!]
\centering
\includegraphics[width=1.0\linewidth]{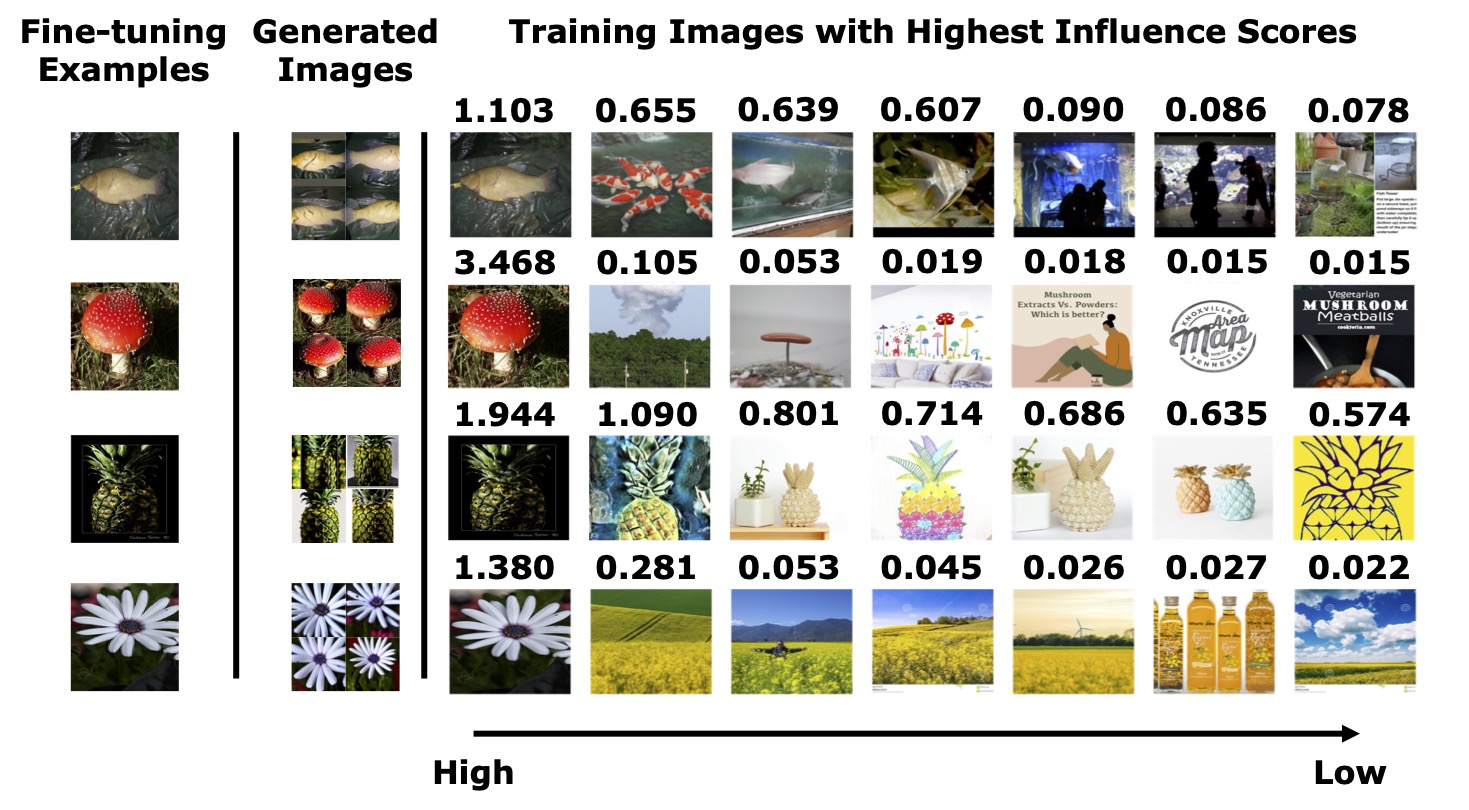}
\caption{\textbf{Data attribution in diffusion models.} For given synthesized samples of the second column, obtained by fine-tuning with an image of the first column, we illustrate the points with the highest influences in the candidate set. Our method can assign the highest influence to the fine-tuning point which computationally influences the synthesized image the most.
}
\vspace{-0.5em}
\label{fig:diff}
\vspace{-1em}
\end{figure}

\noindent\textbf{Motivation.} Recent advancements in generative models, such as stable diffusion~\citep{Rombach_2022_CVPR}, have shown remarkable performance in synthesizing high-quality images. Even though the generated images differ from the original training data, these are largely influenced by them, raising potential issues of copyright infringement~\citep{wang2023evaluating}. Therefore, it is important to identify the training points contributing most to synthesizing a specific output. 


\noindent\textbf{Setup.} Following~\citep{wang2023data}, we leverage the stable diffusion~\citep{Rombach_2022_CVPR} as a pre-trained model and a specific concept in ImageNet with various pre-defined prompts related to the concept for fine-tuning. This will generate a model customized to the concept. The synthesized images from the customized model are computationally influenced by the fine-tuning examples by construction. Hence, the fine-tuning examples can be regarded as \emph{ground-truth} high-influence training points. Our goal is to test whether the proposed method can indeed assign a higher influence to the fine-tuning points than the points in the pre-training set. 
We regard each individual synthesized image as $D_\text{tst}$ and apply our method to calculate the influence of individual training points in both the pre-training set (i.e., LAION 2B~\citep{schuhmann2022laion}) and the fine-tuning set. We adopt a strategy similar to~\citep{akyurek2022tracing} to accelerate the evaluation. In particular,
we create a candidate set $D_\text{candidate}\subset D_\text{trn}$~\citep{akyurek2022tracing}, consisting of: (1) Ground truth, which is the fine-tuning examples and (2) Distractors, which include $50$ pre-training samples whose captions have the most overlap with the given prompt as well as $50$ random samples drawn from the pre-training set. 

\noindent\textbf{Result.} Figure~\ref{fig:diff} shows the qualitative results of our proposed method. We retrieved a set of 7 samples with the highest influence scores from the candidate set.
We observe that the fine-tuning image is consistently ranked with the highest influence. This aligns with the design of our experiment, where the fine-tuned example is deliberately constructed to exert the most computational influence on the synthesized image. Also, it can be seen that
the remaining samples retrieved mostly share similarities either in image or caption or both with the synthesized image. 
We also provide the quantitative results in identifying the ground truth across different sizes of candidate sets and the comparison with the baselines in the Appendix~\ref{app:analysis and details}. 

\subsection{Data Leakage Detection}
\label{sec: training data leakage subsection}
\begin{table}[t!]
  \centering
    \renewcommand{\arraystretch}{2.0}
    \scalebox{0.74}{
        \begin{tabular}{lrrrrrrrr}
            & \multicolumn{4}{c}{\textbf{CIFAR-10 -- ResNet18}} & \multicolumn{4}{c}{\textbf{CIFAR-100 -- ResNet50}} \\ \cmidrule[0.4pt](l{0.525em}r{0.525em}){2-5}  \cmidrule[0.4pt](l{0.525em}r{0.525em}){6-9}
           \textbf{Method}  & \textbf{Time} & \textbf{T-1}       & \textbf{T-5}       & \textbf{T-10}      & \textbf{Time} & \textbf{T-1}       & \textbf{T-5}       & \textbf{T-100}       \\ \cmidrule[0.4pt](l{0.525em}r{0.525em}){1-1}  \cmidrule[0.4pt](l{0.525em}r{0.525em}){2-2} \cmidrule[0.4pt](l{0.525em}r{0.525em}){3-3} \cmidrule[0.4pt](l{0.525em}r{0.525em}){4-4} \cmidrule[0.4pt](l{0.525em}r{0.525em}){5-5} \cmidrule[0.4pt](l{0.525em}r{0.525em}){6-6} \cmidrule[0.4pt](l{0.525em}r{0.525em}){7-7} \cmidrule[0.4pt](l{0.525em}r{0.525em}){8-8} \cmidrule[0.4pt](l{0.525em}r{0.525em}){9-9}
\texttt{IF-100}     & 10 & 30     & 35     & 45  & 38   & 0     & 0     & 20         \\
\texttt{IF-1000 }   & 15 & 45     & 45     & 50  & 92   & 0     & 0     & 25          \\
\texttt{IF-10000 }  & 72 & 0     & 0     & 0  & 570   & 0     & 0     & 0         \\
\texttt{TracIn-1 }  & 12 & 2     & 6     & 7  & 43   & 6     & 11    & 19         \\
\texttt{TracIn-3 }  & 36 & 6     & 10    & 14 & 130  & 9     & 14    & 20          \\
\texttt{TracIn-5 }  & 60 & 6     & 10    & 14 & 215  & 10    & 13    & 19          \\
\textbf{\texttt{Forward-INF}} & \textbf{0.3} & \textbf{100}    & \textbf{100}     & \textbf{100}     & \textbf{1.5}     & \textbf{95}      & \textbf{100}           & \textbf{100}   \\
 \cmidrule[0.9pt](l{0.525em}r{0.525em}){1-1} \cmidrule[0.9pt](l{0.525em}r{0.525em}){2-2} \cmidrule[0.9pt](l{0.525em}r{0.525em}){3-3} \cmidrule[0.9pt](l{0.525em}r{0.525em}){4-4} \cmidrule[0.9pt](l{0.525em}r{0.525em}){5-5} \cmidrule[0.9pt](l{0.525em}r{0.525em}){6-6} \cmidrule[0.9pt](l{0.525em}r{0.525em}){7-7} \cmidrule[0.9pt](l{0.525em}r{0.525em}){8-8} \cmidrule[0.9pt](l{0.525em}r{0.525em}){9-9}
\end{tabular}
}
\caption{\textbf{Data leakage detection} with top-K (T-K) accuracy (\%). Performance is compared on different datasets and models. Here, \texttt{TracIn-K} denotes that we use K checkpoints for \texttt{TracIn} and \texttt{IF-L} denotes that we used L depths for \texttt{IF}. We also report computation time (in minutes) for each test point.} 
\label{tab:train-test leakage}
\vspace{-1em}
\end{table}


\noindent\textbf{Motivation.} Data leakage refers to an oftentimes unintended mistake that is made by the creator of a machine learning model in which they accidentally share the information between the test and training data set. In this task, we apply data influence estimation methods to detect data leakage. Intuitively, for a given test point, if there exists a leakage to the training set, then the corresponding leaked duplicate point would have the highest influence. 
To evaluate the detection performance of the leaked samples, we use the top-k detection rate metric.  

\noindent\textbf{Setup.} In training data leakage evaluation, we use ResNet-18 (RN18) and ResNet-50 (RN50) classifiers~\citep{he2016deep}, trained on CIFAR-10~\citep{krizhevsky2009learning} and CIFAR-100~\citep{deng2009imagenet}, respectively. To simulate the case of data leakage, we set $D_{tst} \subset D_{trn}$. For baseline methods, we vary the key hyperparameters (i.e., the number of depths for Taylor expansion used by LISSA in approximating the Hessian for \texttt{IF}~\citep{agarwal2017second, park2023trak} and the number of checkpoints for \texttt{TracIn}). We present the details and results of ImageNet100 in Appendix~\ref{app:analysis and details}.

\noindent\textbf{Result.} As shown in Table~\ref{tab:train-test leakage}, our approach is effective in identifying duplicated samples with 100\% and 95\% top-1 detection accuracy for both RN18 and RN50 classifiers. 
We observe that with the larger model RN50, the performance of \texttt{IF} deteriorates when detecting data leakage. This result is consistent with~\citep{basu2020influence} that \texttt{IF}s are poor estimates of the impact of excluding a training point for neural networks. 
Regarding the depth parameter, the algorithm's convergence to  $\alpha^{-1}(G + \lambda I)^{-1}v$ depends on the condition $\alpha(\tilde{G} + \lambda I) \preceq I$ being valid at every step ~\citep{grosse2023studying}, which is rarely the case with large and complex models. Hence, errors could accumulate with each iteration, leading to the worst result for the \texttt{IF-10000}. Moreover using more iterations can result in a larger per-iteration cost.
It is interesting to note that \texttt{TracIn} also exhibits poor performance in data leakage detection. 
Although it often identifies visually similar samples to the test example, it fails to accurately detect the ground-truth leaked sample as illustrated in Figure ~\ref{fig:IF_TracIn}. This discrepancy arises because, although the gradient of a duplicated sample may align with that of a test sample, numerous other training samples may align in a similar direction but with greater magnitude, yielding even higher scores than the duplicate itself.
Additionally, \texttt{TracIn} in large-scale models often relies on last-layer gradient information ~\citep{akyurek2022tracing}, which may lead to a suboptimal approximation to ground truth attribution.
As shown in Table~\ref{tab:train-test leakage}, \algname achieves the best scores in terms of both efficacy and computational efficiency, while other approaches suffer from a significant computational bottleneck (benchmarked on NVIDIA GeForce RTX 2080 Ti), as well as difficulty in finding duplicated pairs. 

\subsection{Memorization Analysis}
\label{sec: memorization}
\begin{figure}[t!]
\centering
\includegraphics[width=0.8\linewidth]{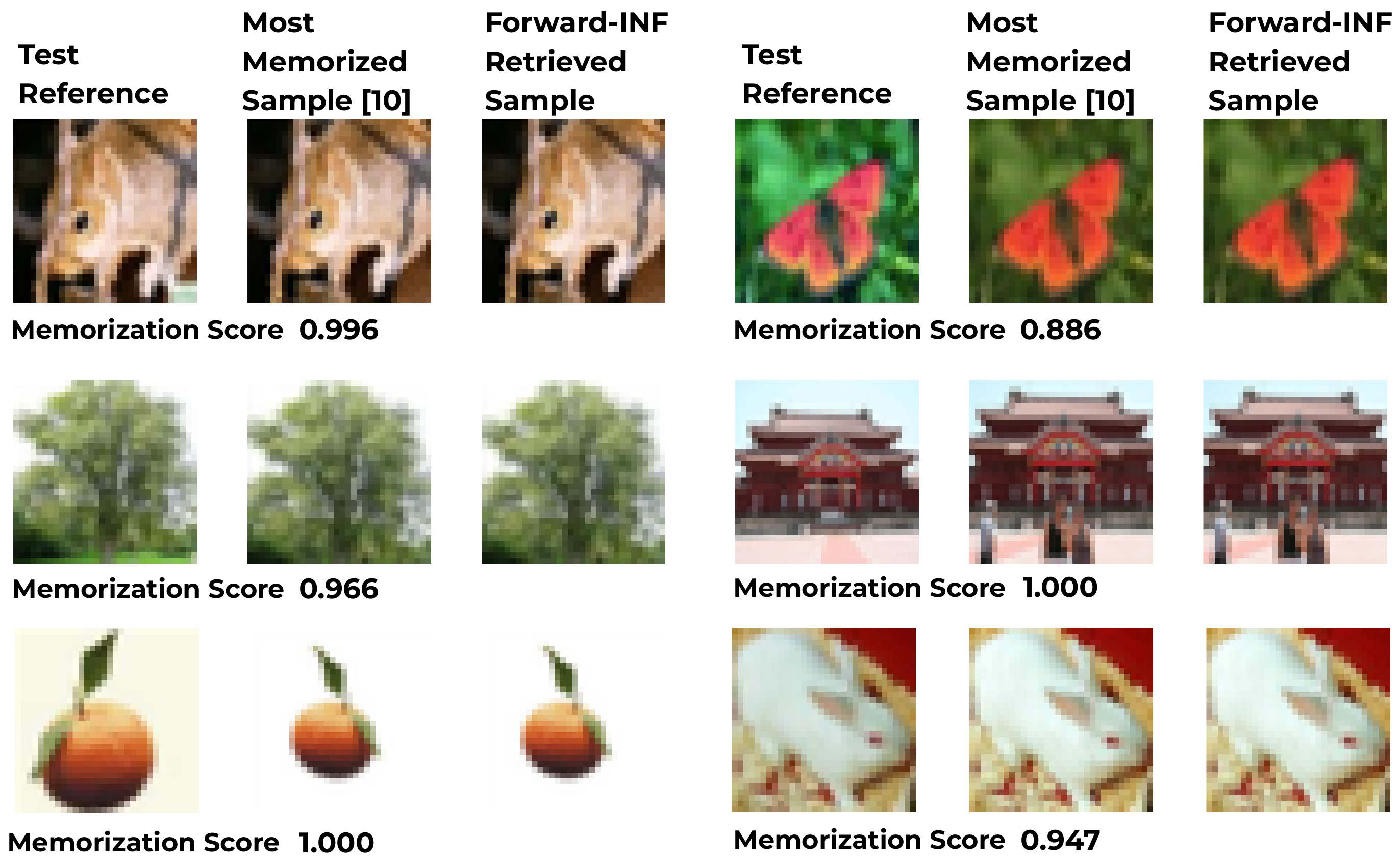}
\caption{\textbf{Memorization analysis}, where the goal is to identify which training point's memorization is critical for predicting a specific test point. Prior work~\citep{feldman2020neural} proposed an algorithm to compute memorized training-test pairs, but it requires re-training the target model many times. We show that \algname can identify the same memorized pairs without the need for re-training.   
}
\vspace{-1em}
\label{fig:memorization}
\end{figure}

\noindent\textbf{Motivation.} Prior research~\citep{feldman2020neural}, studied the pair of training and test samples in which memorization of the training point is important to predicting a given test sample. Characterizing these memorized pairs is essential for understanding the learning mechanisms in neural networks. Towards that end, they proposed a re-training-based approach that directly evaluates the train-to-test influence in order to identify such so-called ``influential pairs'' (following their terminology). However, this approach entails high computational requirements. Here, we aim to examine whether our approach, which is much more light-weighted than~\citep{feldman2020neural}, can identify the same pairs. 

\noindent\textbf{Setup.} To ensure a fair comparison, we employ the same training algorithm as described in ~\citep{feldman2020neural} to train ResNet-50 classifiers on CIFAR-100. We utilize the influential pairs provided by the authors for a qualitative comparison. 

\noindent\textbf{Results.} In Figure~\ref{fig:memorization}, we compare the most influential training point as determined by our influence score calculations with the results from prior research. In each sub-figure, the first two panels display the test point and the most influential training point characterized by~\citep{feldman2020neural}, while the third one presents the top-influence point retrieved by \algname. Figure~\ref{fig:memorization} shows that our approach can identify the same influential pairs as those in~\citep{feldman2020neural} but without retraining models. As mentioned in~\citep{feldman2020neural}, the high-influential pairs (with an influence score $>0.4$) are near duplicated samples and benefit the most from memorization. Therefore, it leads to high memorization scores. Further qualitative results are provided in Appendix~\ref{app:analysis and details}.

\subsection{Mislabeled Data Detection}
\label{sec: mislabeled detection}

\noindent\textbf{Motivation.} Automated identification of incorrectly labeled samples in training datasets is essential, particularly given the high incidence of human labeling errors~\citep{karimi2020deep}. Human judgment often varies and can be subjective, leading to inconsistent labeling, a situation that is particularly pronounced in the case of ambiguous samples. Reliable mislabeled data detection can substantially reduce the costs associated with human labeling by facilitating automated checks.

\noindent\textbf{Setup.} For our study, we select a random training subset $D_{trn}$ of 2000 data points from the CIFAR-10 dataset, intentionally introducing label errors in 20\% of these samples by assigning them to random classes. This data size is used due to the computational complexities associated with the Influence Function (\texttt{IF}).  We then train a ResNet-18 model for 100 epochs. 
We follow the same setting by calculating self-influence scores, i.e. the influence of the training point onto itself, without relying on the validation data, i.e., $D_\text{tst} = D_\text{trn}$. After computing the scores for each training point, we sort them in descending order and then examine them for potential mislabeling in this sorted order.

\begin{figure}[h!]
\centering
\includegraphics[width=0.7\linewidth]{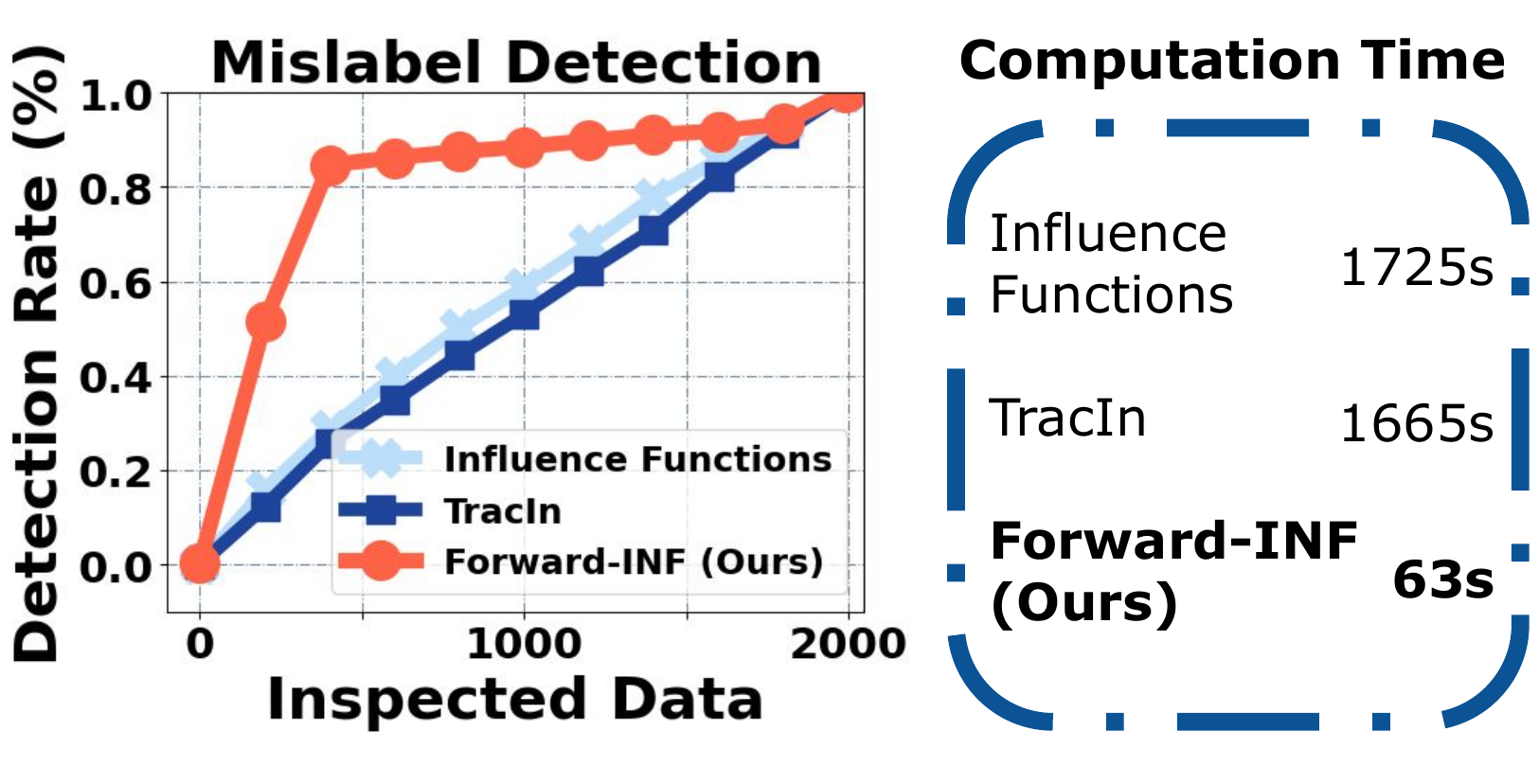}
\caption{\textbf{Mislabeled data detection in a subset of CIFAR-10}. Left) Mislabeled data detection performance comparison between \algname and \texttt{IF}. Right) Computation time comparison between methods. \algname is not only effective in detecting mislabeled training data but also efficient in its computation.}


\label{fig:mislabeled detection}
\vspace{-1em}
\end{figure}

\noindent\textbf{Result.} In Figure ~\ref{fig:mislabeled detection}, we show the mislabeled data detection result. As shown in the figure, our proposed method can find over 80\% of mislabeled samples within the first 300 checked samples, whereas \texttt{IF} would need to go through 75\% of training samples to detect that many mislabeled samples, regardless of the sorting order. For the case of computing self-influence, the gradient-based methods, \texttt{IF} and \texttt{TracIn}, in fact, compute the magnitude of each point's gradient. Thus, even though the gradients of mislabeled training points might point in the opposite direction than the gradients of the clean samples, their magnitude can be smaller than those of the clean ones, resulting in the incapability of successful mislabel detection. Our method overcomes this issue by computing the loss difference between two models, $\hat{\theta}$ and $\hat{\theta}_{+D_{tst}}$, where the loss difference for clean samples will be smaller than those of the mislabeled ones, since clean samples highly likely have samples with similar labeling distribution, while the randomly mislabeled samples hardly have the support in the training dataset. Thus, they are more prone to larger loss changes. In conclusion, our approach demonstrates high mislabeled data detection performance and computation efficiency compared to existing baselines. 


\subsection{Language Model Behavior Tracing}
\label{sec: model behavior tracing}
\noindent\textbf{Motivation.} With the growing prevalence of large language models in various applications, such as conversational agents~\citep{thoppilan2022lamda,shuster2022blenderbot,meta2022human}, the importance of providing reasonable supporting evidence has become paramount. As a result, the need to trace the origin of a model's output back to specific data samples has also become indispensable to identifying the responsible training data.
Driven by this motivation, we study the task of model behavior tracing regarding factual assertion, which involves identifying the training examples responsible for inducing the model to make some factual assertion at test time.
 
\noindent\textbf{Setup.} We utilize a MT5 model~\citep{xue2020mt5} to finetune on the FTRACE-TREx dataset ~\citep{akyurek2022tracing}. We consider each training example that conveys the same fact as a ``proponent'' of the corresponding test example and as a ``distractor'' otherwise. We provide the details of the dataset and hyperparameter selection in Appendix~\ref{app:hyperparameter setting}. Due to the demanding computation required by other direct train-to-test influence estimation methods, such as \texttt{IF}, here we only compare with \texttt{TracIn}.
In order to make \texttt{TracIn}~\citep{pruthi2020estimating} more efficient, for each test sample, we limit the scope of our search to a \emph{candidate set}, i.e., a subset of the entire training dataset, following previous studies ~\citep{akyurek2022tracing, park2023trak}. 
We leverage the same evaluation metrics (i.e., precision and Mean Reciprocal Rank (MRR)) described in the previous studies ~\citep{akyurek2022tracing, park2023trak}. To consider the efficiency and the performance simultaneously, we propose a time-dependent performance metric, i.e., performance in a limited time budget. This metric is realistic because, in practice, user-facing products cannot afford to spend an indefinite amount of time responding to a user's request.

\begin{table}[h]
\centering
\renewcommand{\arraystretch}{1.9}
\resizebox{\linewidth}{!}{
\begin{tabular}{lccccr}
\textbf{Candidate Set Size}               & \multicolumn{2}{c}{\textbf{15K}}           & \multicolumn{2}{c}{\textbf{20K}}           & \textbf{Inspected Queries} \\ 
\cmidrule[0.4pt](l{0.525em}r{0.525em}){1-1} \cmidrule[0.4pt](l{0.525em}r{0.525em}){2-3} \cmidrule[0.4pt](l{0.525em}r{0.525em}){4-5} \cmidrule[0.4pt](l{0.525em}r{0.525em}){6-6}
\textbf{Metric}         & \textbf{MRR}    & \textbf{Precision} & \textbf{MRR}    & \textbf{Precision} & \textbf{\# of Queries/Min} \\ 
\cmidrule[0.4pt](l{0.525em}r{0.525em}){1-1} \cmidrule[0.4pt](l{0.525em}r{0.525em}){2-2} \cmidrule[0.4pt](l{0.525em}r{0.525em}){3-3} \cmidrule[0.4pt](l{0.525em}r{0.525em}){4-4} \cmidrule[0.4pt](l{0.525em}r{0.525em}){5-5} \cmidrule[0.4pt](l{0.525em}r{0.525em}){6-6} 
\texttt{TracIn} (Single) & 0.1658 & 0.1367    & 0.1532 & 0.1300    & 307.522           \\
\texttt{TracIn} (Multi)  & 0.1596 & 0.1300    & 0.1508 & 0.1300    & 307.522           \\
\textbf{\algname} & \textbf{0.2101} & \textbf{0.1650} & \textbf{0.1927} & \textbf{0.1518} & \textbf{1306.323} \\ 
\cmidrule[0.9pt](l{0.525em}r{0.525em}){1-1} \cmidrule[0.9pt](l{0.525em}r{0.525em}){2-2} \cmidrule[0.9pt](l{0.525em}r{0.525em}){3-3} \cmidrule[0.9pt](l{0.525em}r{0.525em}){4-4} \cmidrule[0.9pt](l{0.525em}r{0.525em}){5-5} \cmidrule[0.9pt](l{0.525em}r{0.525em}){6-6}
\end{tabular}
}
\caption{\textbf{Language model behavior tracing} performance comparison of different attribution methods.}
\label{tab:behavior tracing}
\vspace{-1em}
\end{table}

\noindent\textbf{Results.} As shown in Table~\ref{tab:behavior tracing}, we observe that \algname outperforms the \texttt{TracIn} in terms of both metrics as \texttt{TracIn} cannot inspect enough samples within the given time. 
Also, counter-intuitively, the performance for \texttt{TracIn} drops when using multiple checkpoints compared with a single checkpoint. This finding is also reported in the previous studies~\citep{akyurek2022tracing, park2023trak}.
In addition, \algname is more than four times faster than \texttt{TracIn}, in terms of the inspected number of queries per minute, even if we calculate only one layer's gradient for \texttt{TracIn}. Therefore, in the domain of large-scale models trained on vast quantities of data samples, the benefit of our method stands out. We further provide behavior-tracing experiments on paraphrased queries in Appendix~\ref{app:analysis and details}. We also provide a comparison with a simple model-independent information retrieval~\citep{robertson1995okapi} approach in Appendix~\ref{app:analysis and details}.

\section{Conclusion}

Our contribution lies in the investigation of the Mirrored Influence Hypothesis. Expanding upon this hypothesis, we have developed a novel method to estimate train-to-test influence by solving the test-to-train influence problem. This approach involves evaluating the impact of incorporating a specific test set into the training set on the prediction of a training data source. Our method can be applied broadly and contribute meaningful insights across various settings. In particular, it outperforms traditional approaches that directly compute train-to-test influence by achieving an improved tradeoff between utility and efficiency.

\noindent\textbf{Limitations \& Future Work.} The exploration of the Hypothesis unveils many avenues for future research. Although this paper excludes heuristic enhancements to \algname, it is expected that strategic layer selection, and incorporating strategies to counter catastrophic forgetting~\citep{kirkpatrick2017overcoming}, could enhance our method's performance. Formulating a theoretical framework to formally validate the Hypothesis constitutes an intriguing direction for future studies.



\section{Acknowledgment}
We thank Hoang Anh Just and Himanshu Jahagirdar from the ReDS lab for their invaluable help in experiments and discussion. 
RJ and the ReDS lab acknowledge support through grants from the Amazon-Virginia Tech Initiative for Efficient and Robust Machine Learning, the National Science Foundation under Grant No. IIS-2312794, NSF IIS-2313130, NSF OAC-2239622, and the CCI SWVA Research Engagement Award.

{
    \small
    \bibliographystyle{ieeenat_fullname}
    \bibliography{main}
}

\appendix
\cleardoublepage
\newpage

\section{Related work}
\label{sec:Related}
We refer the readers to a recent survey~\citep{hammoudeh2022training} for a detailed account of research in data influence estimation. In this section, we will focus on key ideas and representative prior works in this field and elaborate on their connections with our work.

\paragraph{Gradient-based influence estimation.} Intuitively, when models are trained with data, each training data has a unique gradient trace. Many studies have been focusing on measuring the influence of data through the lens of gradient alignment between training and test samples. Koh and Liang ~\citep{koh2017understanding} leveraged a classic robust statistics concept, Influence Function ~\citep{cook1982residuals}, to quantify training data influence. 
\texttt{IF} evaluates the effect of infinitesimal change on the loss associated with an individual training point on the test loss. The computation of \texttt{IF} requires inverting the Hessian matrix, which is prohibitively expensive for modern neural networks. To tackle this challenge, Koh and Liang ~\citep{koh2017understanding} proposed to compute \texttt{IF} via iteratively approximating the Hessian-vector product (HVP).
Pruthi et al.~\citep{pruthi2020estimating} presented \texttt{TracIn}, a technique to estimate the influence of each training data by exploiting the gradient over all iterations. In particular, this influence estimator relies on the gradient of a test loss and a training loss. To scale up the approach, a practical alternative was proposed that considers a few checkpoints to calculate the gradient rather than using full iterations to approximate the data influence. Our work studies the duality associated with these gradient-based influence quantification schemes and leverages the duality to propose a more efficient alternative that does not require calculating gradients for individual training points.

\textbf{Re-training based influence estimation.} Re-training-based methods follow a general recipe that starts by training models on different training data subsets and then examining how the performance of these models changes when a given training point is added to the subsets. Ilyas et al. introduced Datamodel ~\citep{ilyas2022datamodels} which involves training thousands of models to estimate the data influence of each training datum. Specifically, this method leverages a parameterized surrogate model to predict the model performance based on the input training set and the surrogate model is learned from a training set consisting of pairs of an input subset and the corresponding model performance.
Park et al. ~\citep{park2023trak} proposed TRAK, which leveraged several techniques to enhance the efficiency of Datamodel. TRAK linearizes the model output function using Taylor approximation and reduces the dimensionality of the linearized model using random projections. However, it still requires repeated model training.

Another line of research does not train surrogate models for data influence estimation; instead, they directly compute a weighted average of the model performance changes in response to the addition of a training point across different subsets. Notable examples include the Shapley value~\citep{jia2019efficient,ghorbani2019data}, Beta Shapley \citep{kwon2021beta} and Banzhaf value \citep{wang2022data}, which differ in the design of the weighting scheme over different subsets. However, these methods face significant computational challenges for large models due to the need of retraining models on different subsets.
Just et al.~\citep{just2023lava} recently proposed LAVA as a scalable solution that evaluates data influence on the model performance using optimal transport. Despite LAVA's efficiency, LAVA does not provide a way for monitoring the training data's contribution to the model prediction on the individual test point.

\begin{figure}[ht!]
\centering
\includegraphics[width=\linewidth]{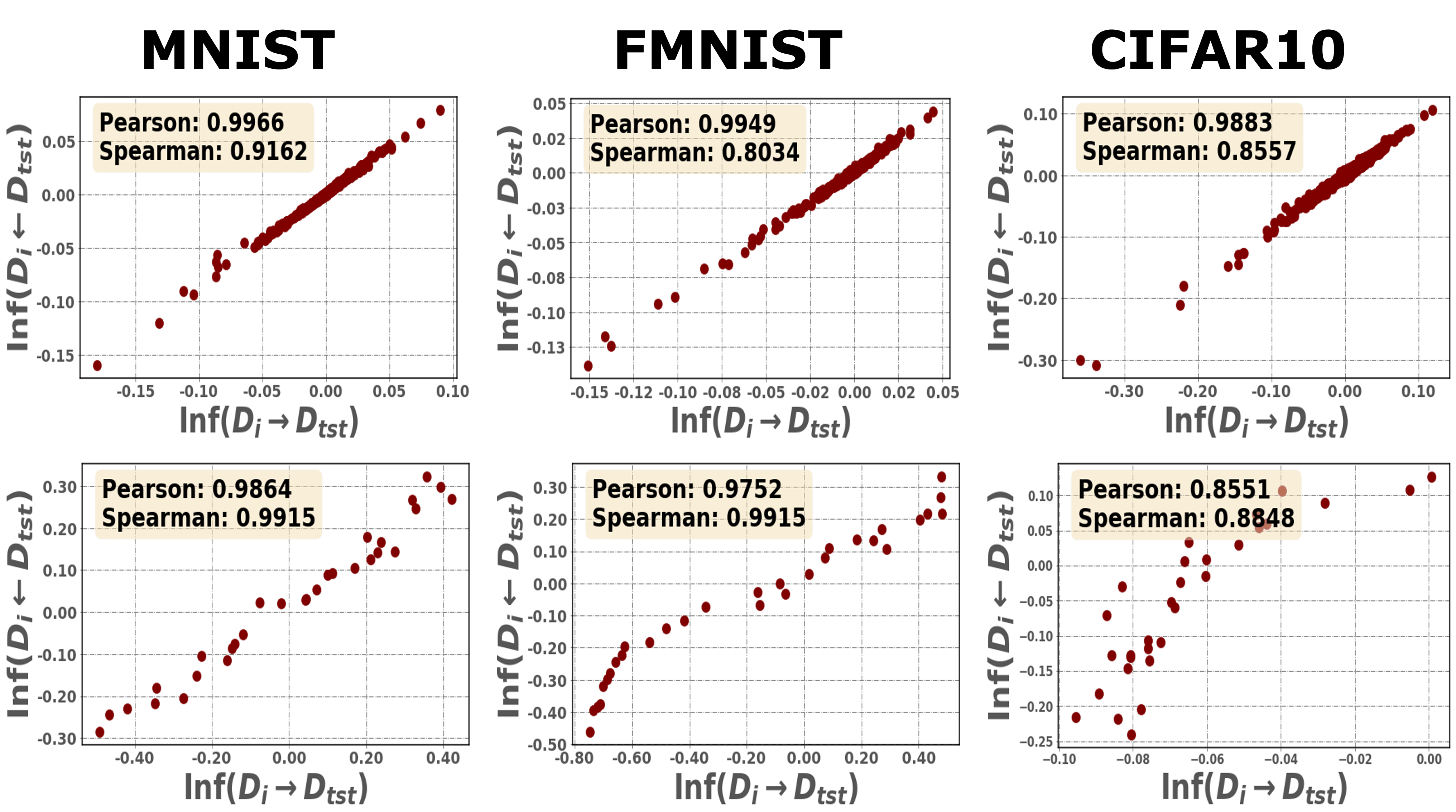}
\caption{Correlation scores across different datasets under the noisy and near-noiseless settings. The first row shows the correlations between $\text{Inf}(D_i\rightarrow D_\text{tst})$ and  $\text{Inf}(D_i\leftarrow D_\text{tst})$ under the near-noiseless setting, and the second row presents the correlation results under the noisy setting. We note that the correlation scores for CIFAR-10 under the noisy setting can be increased to 0.8971, and 0.9551 respectively by changing the number of epochs.}
\vspace{-1em}
\label{fig:Corr on datasets and settings}
\end{figure}

\begin{table}[ht]
\centering
\renewcommand{\arraystretch}{1.5} 
\resizebox{0.8\linewidth}{!}{ 
\begin{tabular}{llcccc}
\multicolumn{2}{l}{\textbf{Validation Set Size}}     & \textbf{50}     & \textbf{100}    & \textbf{1000}   \\
\cmidrule[0.4pt](l{0.525em}r{0.525em}){1-2} \cmidrule[0.4pt](l{0.525em}r{0.525em}){3-3} \cmidrule[0.4pt](l{0.525em}r{0.525em}){4-4} \cmidrule[0.4pt](l{0.525em}r{0.525em}){5-5}
\multirow{2}{*}{\textbf{MNIST}}  & Pearson  & 0.9208 & 0.9512 & 0.9652 \\
                                 & Spearman & 0.9755 & 0.9764 & 0.9942 \\
\multirow{2}{*}{\textbf{FMNIST}} & Pearson  & 0.9244 & 0.9611 & 0.9786 \\
                                 & Spearman & 0.9568 & 0.9795 & 0.9964 \\
\multirow{2}{*}{\textbf{CIFAR10}} & Pearson & 0.6918 & 0.8401 & 0.8551 \\
                                 & Spearman & 0.8274 & 0.9034 & 0.8848 \\
 \cmidrule[0.4pt](l{0.525em}r{0.525em}){1-1}
 \cmidrule[0.4pt](l{0.525em}r{0.525em}){2-2} \cmidrule[0.4pt](l{0.525em}r{0.525em}){3-3} \cmidrule[0.4pt](l{0.525em}r{0.525em}){4-4} \cmidrule[0.4pt](l{0.525em}r{0.525em}){5-5}
\end{tabular}}
\caption{Correlation scores across different sizes of the validation set. We keep the same mislabeled ratio (i.e., 0.5) while changing the sizes of the validation set.}
\label{tab:Corr on diff validation set size}
\end{table}

\begin{table}[ht]
\centering
\renewcommand{\arraystretch}{1.5} 
\resizebox{0.8\linewidth}{!}{ 
\begin{tabular}{llcccc}
  \multicolumn{2}{l}{\textbf{Mislabeled Ratio}}     & \textbf{20\%}    & \textbf{30\%}    & \textbf{50\%}    \\
\cmidrule[0.4pt](l{0.525em}r{0.525em}){1-2} \cmidrule[0.4pt](l{0.525em}r{0.525em}){3-3} \cmidrule[0.4pt](l{0.525em}r{0.525em}){4-4} \cmidrule[0.4pt](l{0.525em}r{0.525em}){5-5}

\multirow{2}{*}{\textbf{MNIST}}  & Pearson  & 0.9621 & 0.9864 & 0.9640 \\
                                 & Spearman & 0.9653 & 0.9915 & 0.9907 \\
\multirow{2}{*}{\textbf{FMNIST}} & Pearson  & 0.9421 & 0.9857 & 0.9752 \\
                                 & Spearman & 0.9479 & 0.9840 & 0.9915 \\
\multirow{2}{*}{\textbf{CIFAR10}} & Pearson & 0.7521 & 0.9697 & 0.8551 \\
                                 & Spearman & 0.7838 & 0.9702 & 0.8848 \\
\cmidrule[0.4pt](l{0.525em}r{0.525em}){1-1}
 \cmidrule[0.4pt](l{0.525em}r{0.525em}){2-2} \cmidrule[0.4pt](l{0.525em}r{0.525em}){3-3} \cmidrule[0.4pt](l{0.525em}r{0.525em}){4-4} \cmidrule[0.4pt](l{0.525em}r{0.525em}){5-5}
\end{tabular}}
\caption{Correlation scores across different mislabeled ratios in $D_\text{trn}$. We keep the same validation set size (i.e., 500) while changing the mislabeled ratios.} 
\label{tab:Corr on diff mislabeled ratio}
\end{table}

\begin{figure*}[ht!]
\centering
\includegraphics[width=1.0\linewidth]{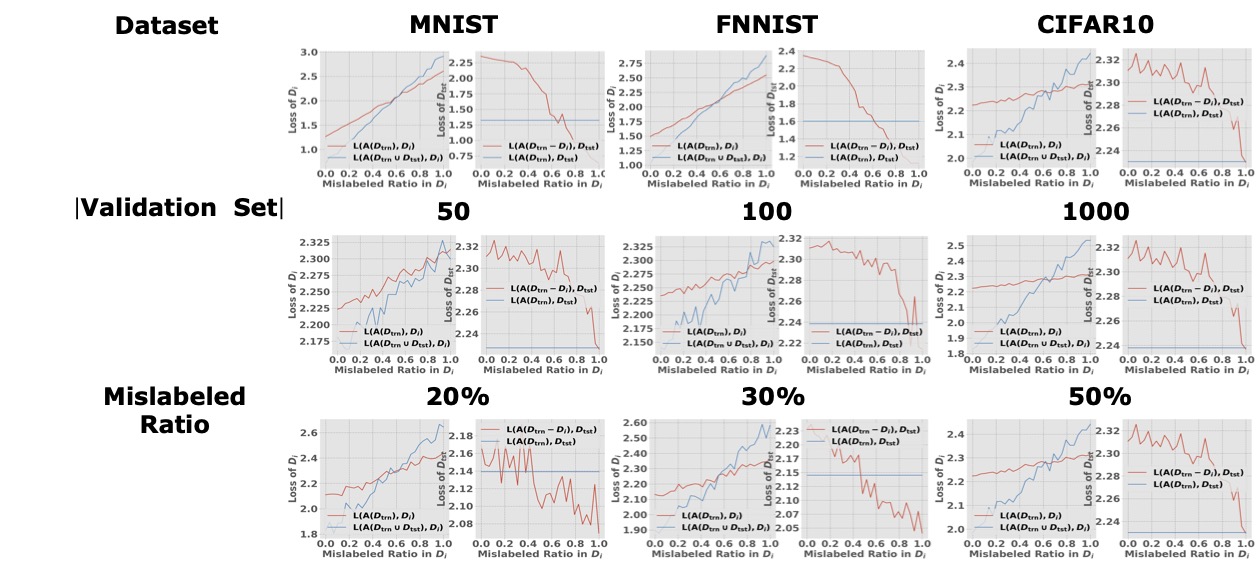}
\caption{Detailed visualization of each value in the calculation of $\text{Inf}(D_i\rightarrow D_\text{tst})$ and $\text{Inf}(D_i\leftarrow D_\text{tst})$ across various datasets, validation set sizes, and mislabeled ratios. The x-axis represents different mislabeled ratios according to different groups, while the y-axis shows the value of each term (i.e.,$\mathcal{L}(\mathcal{A}(D_\text{trn}),D_\text{tst})$, $\mathcal{L}(\mathcal{A}(D_\text{trn}\setminus D_i),D_\text{tst})$, $\mathcal{L}(\mathcal{A}(D_\text{trn}\cup D_\text{tst}),D_i)$, $\mathcal{L}(\mathcal{A}(D_\text{trn}),D_i)$).} 
\label{fig:Corr analysis}
\end{figure*}

\section{Additional Results on Empirical Study of the Mirrored Influence Hypothesis [Section ~\ref{sec: empirical study}]}
\label{app: more empirical results}

In this section, we delve deeper into our Mirrored Influence Hypothesis, as introduced in Section~\ref{sec: empirical study}, by presenting a more detailed analysis and additional results.

\paragraph{Analysis of correlation scores across different datasets.}
We first demonstrate the validity of our hypothesis across different datasets under a near-noiseless setting and a noisy setting.

In the near-noiseless case, we train a logistic regression model using L-BFGS for 1000 iterations with L2 regularization set at 0.001. For the noisy setting, we train a convolutional neural network, consisting of two convolution layers and two MLP layers using SGD with a learning rate of 0.01, a weight decay of 0.001, and a momentum of 0.9. To increase the magnitude of influence of each data point, we select a subset of data, specifically 1050 samples from the MNIST and FMNIST datasets, and 900 samples from the CIFAR-10 dataset. 

As shown in Figure~\ref{fig:Corr on datasets and settings}, we observe the high Pearson and Spearman correlation scores obtained across three datasets and two settings. In the near-noiseless setting, we find high correlation scores across all datasets. In the noisy setting, the CIFAR-10 results show a slight decline in correlation scores. This decrease can be attributed to the limited sample size from CIFAR-10, which potentially led to higher loss and introduced more noise in score calculation.

\paragraph{Analysis of correlation scores across different sizes of the validation set.}

Recall the equations of $\text{Inf}(D_i\rightarrow D_\text{tst})$ and $\text{Inf}(D_i\leftarrow D_\text{tst})$ from Section~\ref{sec:intro}.

\begin{equation}
\resizebox{0.9\columnwidth}{!}{
$\begin{aligned}
    \text{Inf}(D_i\rightarrow D_\text{tst}) = \mathcal{L}(\mathcal{A}(D_\text{trn}),D_\text{tst}) - \mathcal{L}(\mathcal{A}(D_\text{trn}\setminus D_i),D_\text{tst}).
    \nonumber
\end{aligned}$
}
\end{equation}
On the other hand, the \emph{test-to-train} influence characterized by \textbf{P2} can be written as
\begin{equation}
\resizebox{0.9\columnwidth}{!}{
$\begin{aligned}
    \text{Inf}(D_i\leftarrow D_\text{tst}) = \mathcal{L}(\mathcal{A}(D_\text{trn}\cup D_\text{tst}),D_i) - \mathcal{L}(\mathcal{A}(D_\text{trn}),D_i).
       \nonumber
\end{aligned}$
}
\end{equation}

In Section~\ref{sec: empirical study}, when we describe our hypothesis under the noisy setting, we use a group of test points $D_\text{tst}$ to calculate $\text{Inf}(D_i\leftarrow D_\text{tst})$ and $\text{Inf}(D_i\rightarrow D_\text{tst})$ as well as a group of training points (i.e., group-to-group influence). To evaluate the impact of the size of the validation set, we conduct experiments with varying validation set sizes while maintaining the other factors (e.g., mislabeled ratio, and hyperparameters).

Table~\ref{tab:Corr on diff validation set size} shows that enlarging the validation set size (e.g., from 50 to 1000) is advantageous across all datasets. The rationale behind this observation is that a larger validation set (i.e., $|D_\text{tst}|$) yields less sensitivity in the scoring of each term (i.e.,  $\mathcal{L}(\mathcal{A}(D_\text{trn}),D_\text{tst})$ and $\mathcal{L}(\mathcal{A}(D_\text{trn}\setminus D_i),D_\text{tst})$ in $\text{Inf}(D_i\leftarrow D_\text{tst})$. In particular, if we have a larger validation set for $\text{Inf}(D_i\rightarrow D_\text{tst})$, we remove the sensitivity in the choice of validation samples, leading to accurately estimating the effect of each group. Additionally, introducing a larger number of clean samples (i.e., $D_\text{tst}$) to $D_\text{trn}$ might trigger a larger difference between $\mathcal{L}(\mathcal{A}(D_\text{trn}\cup D_\text{tst}),D_i)$ and $\mathcal{L}(\mathcal{A}(D_\text{trn}),D_i)$ due to the amplified negative effect. The second row of Figure~\ref{fig:Corr analysis} further illustrates that enlarging the validation set size enhances the impact of each group. In particular, as depicted by the steeper and smoother blue lines (i.e., $\mathcal{L}(\mathcal{A}(D_\text{trn}),D_\text{tst})$) in the second of the figure, a larger clean validation set helps to mitigate the noise attributable to the stochastic nature of the learning process.

\paragraph{Analysis of correlation scores across different mislabeled ratios.}
In our analysis, we also consider the impact of different mislabeling ratios within the training dataset $(D_\text{trn})$. Table~\ref{tab:Corr on diff mislabeled ratio} underscores the importance of choosing an appropriate mislabeled ratio. This factor is crucial in mitigating stochasticity by amplifying the influence of each group when analyzing correlation scores in a noisy setting. Our empirical study indicates that a mislabeling ratio exceeding 30\% tends to yield less noisy results since it amplifies meaningful signals, such as clear differences between groups. As depicted in the third row of Figure~\ref{fig:Corr analysis}, a 20\% mislabeling ratio is insufficient to reduce noise in score calculation (i.e., more fluctuation in the line of $\mathcal{L}(\mathcal{A}(D_\text{trn} \setminus D_i), D_\text{tst})$).

It is important to avoid excessively high mislabeling ratios, like over 50\%, as they can adversely affect the learning process. For example, with too many mislabeled samples, a model struggles to be effectively trained on the dataset and tends to underfit. This situation makes it difficult to differentiate signals between each group because having many mislabeled samples across different groups may yield a high loss for $\mathcal{L}(\mathcal{A}(D_\text{trn}),D_i)$ and $\mathcal{L}(\mathcal{A}(D_\text{trn}\setminus D_i),D_\text{tst})$ as shown in the third row of Figure~\ref{fig:Corr analysis}. This high loss in the initial stage may not only contain signals of each group's influence but also have additional noise that prevents one from magnifying the influence of each group.

\section{Continual Learning vs. Training from Scratch [Section ~\ref{sec:our approach}]}
\label{app: more theoretical derivations}

As we are considering the new objective of adding a test set $D_\text{tst}$ to the training dataset $D_\text{trn}$, the model trained with the new objective will be $\hat\theta_{\varepsilon,D_\text{tst}}=\arg\min_\theta \mathcal{L}(\theta,D_\text{trn})+\varepsilon\ell(\theta,D_\text{tst})$. Once $\hat\theta_{\varepsilon,D_\text{tst}}$ is obtained, one can evaluate the change in the loss of individual training points:

\begin{equation}
\begin{aligned}
    \text{\algname}(D_i) &= \mathcal{L}(\hat\theta_{\varepsilon,D_{\text{tst}}}, D_i) - \mathcal{L}(\hat\theta, D_i)
\end{aligned}
\end{equation}


The main challenge here is to efficiently obtain $\hat\theta_{\varepsilon,D_\text{tst}}$ from $\hat\theta$. 
Note that for any $\theta$ that is close to $\hat\theta$, we have

\begin{align}
     \mathcal{L}(\theta,D_\text{trn})&+\varepsilon\ell(\theta,D_\text{tst}) \nonumber \\
    = & \, \mathcal{L}(\hat\theta, D_\text{trn})+\varepsilon\ell(\hat{\theta}, D_\text{tst}) \nonumber \\
    & +(\theta-\hat\theta)\cdot\left.\nabla_\theta[\mathcal{L}(\theta,D_\text{trn})+\varepsilon\ell(\theta,D_\text{tst})]\right|_{\theta=\hat\theta} \nonumber \\
    & + \mathcal{O}(\|\theta-\hat\theta\|^2_2) \nonumber \\
    \approx & \, \mathcal{L}(\hat\theta, D_\text{trn})+\varepsilon\ell(\hat\theta,D_\text{tst}) \nonumber \\
    & +(\theta-\hat\theta)\cdot\left.[\nabla_\theta\mathcal{L}(\theta,D_\text{trn})+\varepsilon \nabla_\theta\ell(\theta,D_\text{tst})]\right|_{\theta=\hat\theta} \label{eqn:taylor} \\
    \approx & \, \mathcal{L}(\hat\theta, D_\text{trn})+\varepsilon\ell(\hat{\theta}, D_\text{tst}) \nonumber \\
    & +(\theta-\hat\theta)\cdot\varepsilon \left.\nabla_\theta\ell(\theta,D_\text{tst})\right|_{\theta=\hat\theta}
\end{align}

where the first approximation holds for $\varepsilon\rightarrow0$ and \textit{continuity} of the loss function and the second approximation holds for $\left.\nabla_\theta\mathcal{L}(\theta,D_\text{trn})\right|_{\theta=\hat\theta}=0$ for $\hat\theta$ being the minimizer of $\ell(\theta,D_\text{trn})$ by definition. Taking the $\arg\min$ on both sides of \eqref{eqn:taylor} yields 

\begin{align}
    \hat\theta_{\varepsilon,D_\text{tst}} & = \arg\min_\theta [\mathcal{L}(\theta,D_\text{trn})+\varepsilon\ell(\theta,D_\text{tst})] \nonumber \\
    & \approx \arg\min_\theta \Big[ \mathcal{L}(\hat\theta, D_\text{trn})+\varepsilon\ell(\hat\theta, D_\text{tst}) \nonumber \\
    & \quad +(\theta-\hat\theta)\cdot\varepsilon \left.\nabla_\theta\ell(\theta,D_\text{tst})\right|_{\theta=\hat\theta} \Big] \label{eqn:approx_obj} \\
    & = \arg\min_\theta (\theta-\hat\theta)\cdot\varepsilon \left.\nabla_\theta\ell(\theta,D_\text{tst})\right|_{\theta=\hat\theta}
\end{align}

Note that to find the minimum of \eqref{eqn:approx_obj}, one needs to search along $\nabla_\theta\ell(\theta,D_\text{tst})|_{\theta=\hat\theta}$, i.e., the current gradient direction at the test sample. 
Thus, to find $\hat{\theta}_{\varepsilon, D_\text{tst}}$, given that $\epsilon\rightarrow 0$ and $(\hat{\theta}_{\varepsilon, D_\text{tst}}-\hat\theta)$ is small assuming \textit{continuity} of the loss function, \emph{one only needs to continually update the trained model $\hat\theta$ on the given test sample $D_\text{tst}$}.

In standard practice, $\varepsilon$ is often considered to be positive as it represents the model being \textit{trained} on the test sample (i.e., $\varepsilon\rightarrow 0^+$). 
However, if the test sample is drawn from a similar distribution as the training data (i.e., $\left.\nabla_\theta\ell(\theta,D_\text{tst})\right|_{\theta=\hat\theta}$), the magnitude of the gradient is small.
Interestingly, and counter-intuitively, we can estimate data influence effectively with impressive accuracy by setting $\varepsilon$ to be negative and employing the gradient ascent on the test sample. Namely, we define the following mirrored metric

\begin{equation}
\text{Forward-INF}(D_i) :=\left.\frac{\ell(\hat\theta_{\varepsilon,D_\text{tst}}, D_i) - \ell(\hat\theta, D_i)}{\varepsilon}\right|_{\varepsilon\rightarrow 0^-}
\end{equation}

This technique successfully circumvents numerical issues from diminished gradients on well-trained models and remarkably enhances the accuracy of influence estimation.

\section{Hyperparameter Details [Section ~\ref{sec:experiment}]}
\label{app:hyperparameter setting}

As our approach is based on a gradient ascent (i.e., maximization), selecting appropriate hyperparameters (e.g., learning rate, the number of epochs) is essential. We note that the details of the experiment setting for Section~\ref{sec: empirical study} are elaborated in Section~\ref{app: more empirical results}. In this section, we mainly focus on presenting the details of application experiments from Section~\ref{sec:experiment}.

\paragraph{Data influence estimation in diffusion models [Section~\ref{sec: diffusion}].} In this experiment, it is necessary to fine-tune a stable diffusion pre-trained model on a set of test samples. To fine-tune the pre-trained stable diffusion model, we follow the same setting as that from the previous work ~\citep{wang2023evaluating}. In particular, we randomly pick fine-tuning exemplars from ImageNet~\citep{deng2009imagenet} and further train the pre-trained stable diffusion model on the selected exemplars with 5 iterations, a learning rate of 1e-5, and a batch size of 4. After generating the synthesized samples, we perform gradient ascent on a set of synthesized samples with the same learning rate and 3 iterations. 

\paragraph{Data leakage detection [Section~\ref{sec: training data leakage subsection}].}
We train both ResNet-18 and ResNet-50 using the Adam optimizer with a learning rate of 0.01 and 200 iterations for both CIFAR-10 and CIFAR-100~\citep{krizhevsky2009learning}. 
Due to the computational complexity of our baselines ( \texttt{IF}, \texttt{TracIN} ), we randomly selected 20 test samples for evaluation.
Within this study, we first assign 20\% of the total test samples specifically for the purpose of hyperparameter selection, while the remaining portion is dedicated to evaluation. 

\paragraph{Model behavior tracing [section~\ref{sec: model behavior tracing}].}
In this study, we leverage the FTRACE-TREX dataset ~\citep{akyurek2022tracing} for the model behavior tracing task.
The FTRACE-TREX dataset consists of a set of ``abstracts'' and a set of ``queries'', and each query is annotated with the corresponding list of fact samples ~\citep{akyurek2022tracing}. 
The training set of FTRACE-TREX is sourced from the TREX dataset ~\citep{elsahar2018t}, and the test set of the FTRACE-TREx dataset is derived from the LAMA dataset ~\citep{petroni2019language}. 
Every training example that conveys the identical information as the given test example is designated as a "proponent."
We randomly sampled 100 data points for this task and took an average of three repeated experiments.

\section{Further Analysis and Details} \label{app:analysis and details}

\begin{table}
    \centering
    \renewcommand{\arraystretch}{1.5} 
    \resizebox{0.8\linewidth}{!}{ 
        \begin{tabular}{cccc}
        \textbf{Model/Dataset} & \textbf{Metric} & \textbf{\texttt{IF-100}} & \textbf{\algname} \\
        \cmidrule[0.4pt](l{0.525em}r{0.525em}){1-1} \cmidrule[0.4pt](l{0.525em}r{0.525em}){2-2} \cmidrule[0.4pt](l{0.525em}r{0.525em}){3-3} \cmidrule[0.4pt](l{0.525em}r{0.525em}){4-4}
        ResNet-18 & T-1 & 0.000 & \textbf{0.880} \\
        ImageNet-100 & T-5 & 0.000 & \textbf{0.880} \\
        \cmidrule[0.4pt](l{0.525em}r{0.525em}){1-1} \cmidrule[0.4pt](l{0.525em}r{0.525em}){2-2} \cmidrule[0.4pt](l{0.525em}r{0.525em}){3-3} \cmidrule[0.4pt](l{0.525em}r{0.525em}){4-4}
        \end{tabular}
    }
    \caption{Data Leakage detection performance comparison of ResNet-18 trained on the ImageNet-100 dataset.}
    \label{tab:train-test more leakage}
\end{table}

\begin{figure}[ht!]
\centering
\includegraphics[width=1.0\linewidth]{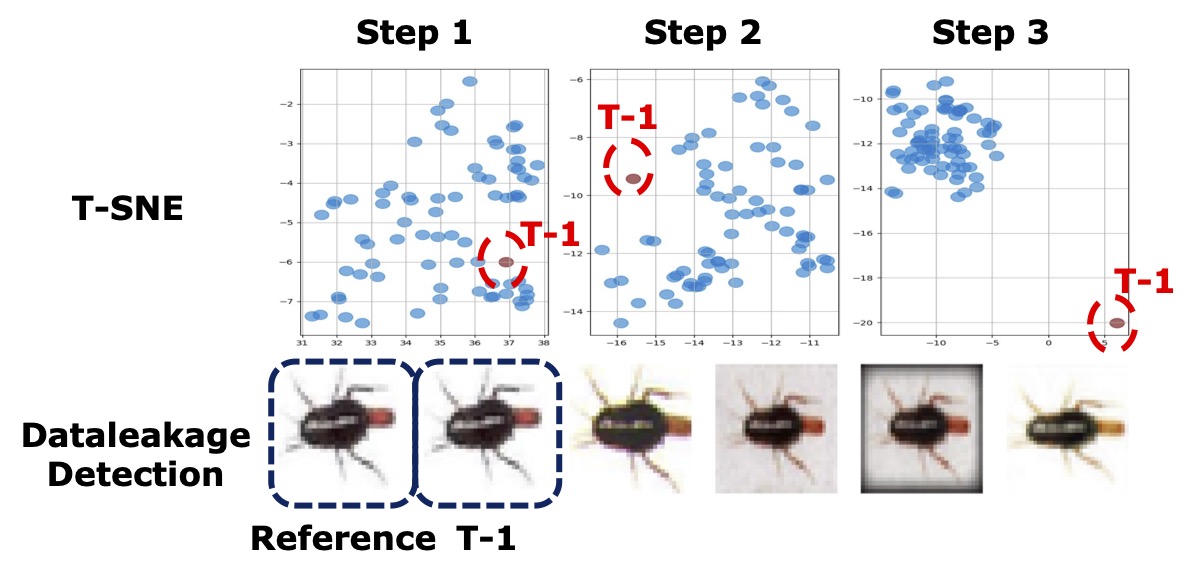}
\caption{Gradient ascent dynamics on T-SNE feature space (top), and top-5 relevant samples retrieved by \algname (bottom). The reference indicates the test sample we want to unlearn and T-1 denotes the top-1 sample retrieved by \algname (bottom) in the T-SNE~\citep{van2008visualizing} feature space (top). We can observe that after a few gradient ascent steps, the top-1 sample begins to migrate away from the center of the feature cluster. This is crucial for effectively detecting duplicated samples.}
\label{fig:unlearning dynamics}
\end{figure}

\subsection{Analysis of Correlation between \algname($D_i$) and $\text{Inf}(D_i\leftarrow D_\text{tst})$}

\begin{table}[ht]
\centering
\renewcommand{\arraystretch}{1.5} 
\resizebox{0.8\linewidth}{!}{ 
\begin{tabular}{llcccc}
\multicolumn{2}{l}{\textbf{Validation Set Size}}       & \textbf{50} & \textbf{100} & \textbf{500} \\
\cmidrule[0.4pt](l{0.525em}r{0.525em}){1-2} \cmidrule[0.4pt](l{0.525em}r{0.525em}){3-3} \cmidrule[0.4pt](l{0.525em}r{0.525em}){4-4} \cmidrule[0.4pt](l{0.525em}r{0.525em}){5-5}
\multirow{2}{*}{\textbf{MNIST}}    & Pearson  & 0.9701 & 0.9819 & 0.9951 \\
                                   & Spearman & 0.9746 & 0.9786 & 0.9942 \\
\multirow{2}{*}{\textbf{FMNIST}}   & Pearson  & 0.9450 & 0.9811 & 0.9942 \\
                                   & Spearman & 0.9479 & 0.9795 & 0.9920 \\
\multirow{2}{*}{\textbf{CIFAR10}}  & Pearson  & 0.9050 & 0.9528 & 0.9862 \\
                                   & Spearman & 0.9052 & 0.9413 & 0.9832 \\
\cmidrule[0.4pt](l{0.525em}r{0.525em}){1-1}
 \cmidrule[0.4pt](l{0.525em}r{0.525em}){2-2} \cmidrule[0.4pt](l{0.525em}r{0.525em}){3-3} \cmidrule[0.4pt](l{0.525em}r{0.525em}){4-4} \cmidrule[0.4pt](l{0.525em}r{0.525em}){5-5}
\end{tabular}}
\caption{Correlation scores between \algname($D_i$) and $\text{Inf}(D_i\leftarrow D_\text{tst})$. We keep the same mislabeled ratio (i.e., 0.5) while changing the sizes of the validation set from 50 to 500.}
\label{tab:Corr on ours vs. dualLOO}
\end{table}

\begin{figure*}[ht!]
\centering
\includegraphics[width=0.9\linewidth]{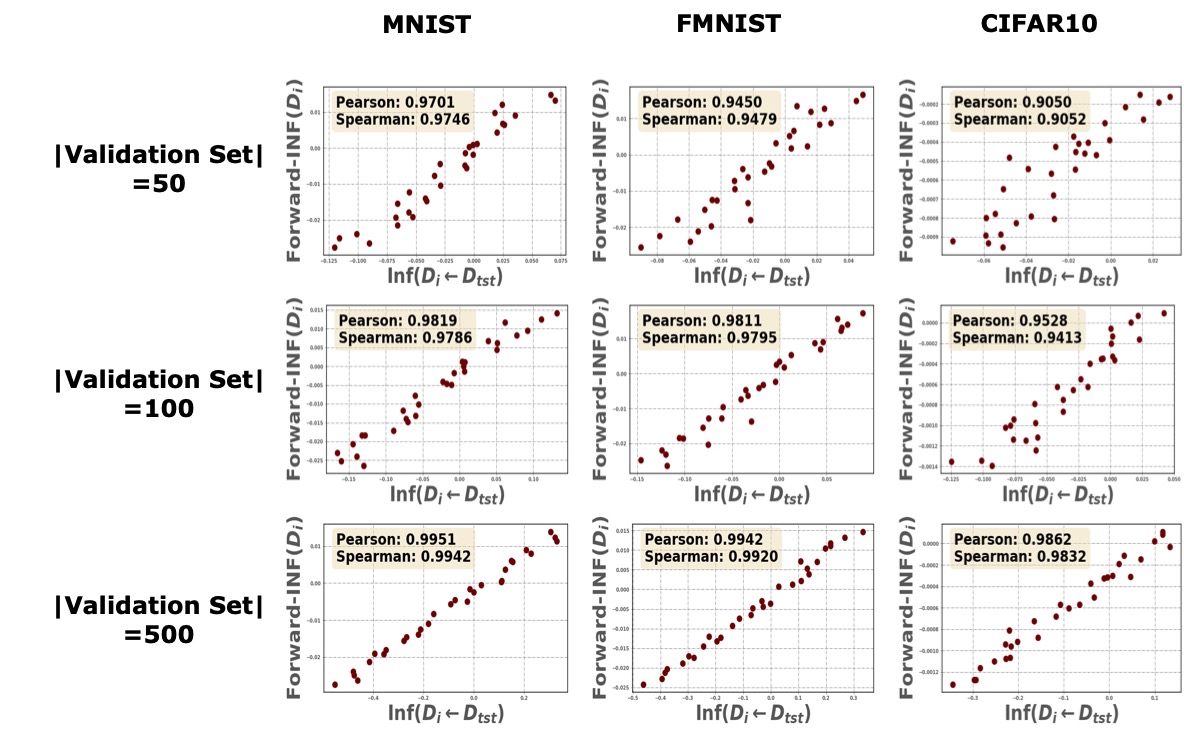}
\caption{Correlation scores between the estimated influence scores from \algname (i.e., \algname($D_i$)) and scores from $\text{Inf}(D_i\leftarrow D_\text{tst})$ across different datasets with different sizes of validation sets.}
\label{fig:Corr between ours and Inf}
\end{figure*}

We want to demonstrate that our influence score approximator, \algname, can effectively estimate the influence score of $\text{Inf}(D_i\leftarrow D_\text{tst})$, thereby obviating the need for re-training since it will be the equivalent influence information as $\text{Inf}(D_i\rightarrow D_\text{tst})$. For this experiment, we chose the noisy setting and utilized the same groups as in previous tests under the noisy setting. We performed gradient ascent with respect to a group of test samples $D_\text{tst}$ and then computed the loss difference between the original and the unlearned models for different groups. We used the SGD optimizer for gradient ascent with a learning rate of 0.01, a weight decay of 0.001, and two iterations. 

Table~\ref{tab:Corr on ours vs. dualLOO} presents the correlation scores between \algname and $\text{Inf}(D_i\leftarrow D_\text{tst})$. As indicated in the table, there is a consistently high correlation between our proposed method and $\text{Inf}(D_i\leftarrow D_\text{tst})$. Similarly, we observe that increasing the size of the validation set results in higher correlation scores, attributed to reduced noise in the learning process. We provide the correlation plots in Figure~\ref{fig:Corr between ours and Inf}.

\subsection{Data Influence Estimation in Diffusion Model  [Section ~\ref{sec: diffusion}]}

In this section, we present both quantitative and qualitative results of data influence estimation for diffusion models. As delineated in ~\citep{wang2023evaluating}, since the fine-tuning examples can serve as the noisy ground truth, we can quantitatively measure whether \algname can identify the ground-truth sample as the most influential sample (Top-1). 
In our experiment, we randomly sample 20 different objects from the ImageNet and also construct the different sizes (e.g., 100, 1K, 10K, 20K) of candidate sets. We follow the same candidate set creation process as we described in Section~\ref{sec: diffusion}. If \algname correctly identifies the ground truth as the most influential point, we mark it as success and average over 20 points. The detection results are presented in Table~\ref{tab:detection_accuracy}.  Our observations reveal that \algname correctly identifies the ground truth fine-tuning examples regardless of its candidate set sizes, demonstrating its robustness across varying candidate set sizes. 

\begin{figure*}[ht!]
\centering
\includegraphics[width=0.90\linewidth]{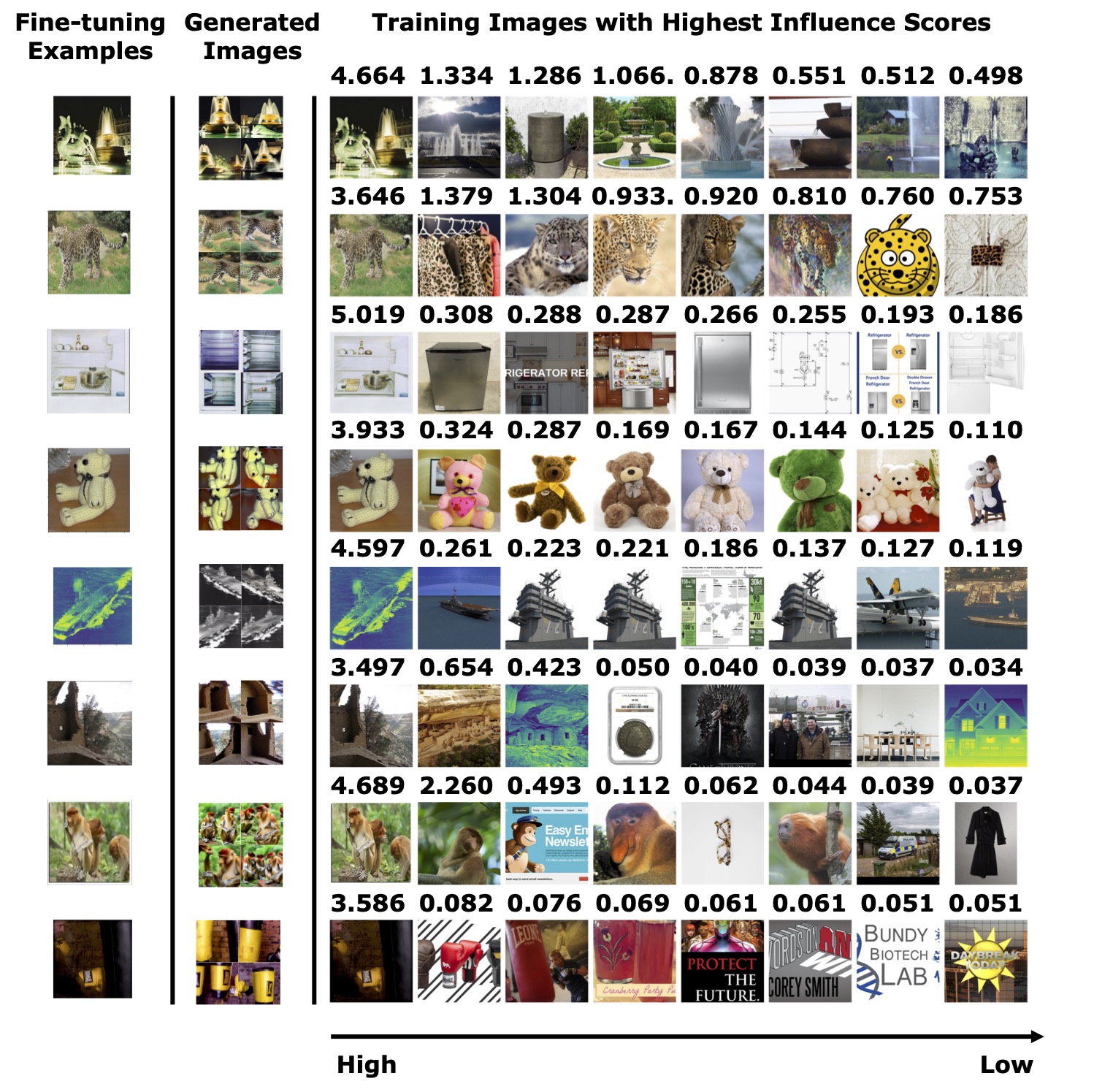}
\caption{Extensive qualitative results of data influence estimation for the diffusion model. Similar to Figure~\ref{fig:diff}, we provide the ground truth (first column), synthesized images after fine-tuning (second column), and training images with the highest influence scores (the remaining columns). \algname can accurately identify the ground truth fine-tuning examples as the most influential data samples contributed to the generation of new images.}
\label{fig:diffusion-extensive}
\end{figure*}

In addition, we provide the qualitative results in Figure~\ref{fig:diffusion-extensive}. In this experiment, we utilize a candidate set comprising 20K samples and present the top 8 highest influential training samples retrieved by \algname. Most of the retrieved samples share similar features with each other either on an image or caption side. This indicates that the unlearning process mostly affects the samples that share similar features. 
Given that the candidate set is curated selectively rather than utilizing the entire dataset, there's a possibility that it might not include more than top-k captions precisely identical to those of the ground truth samples. In this case, \algname might yield some samples that are not directly relevant. However, the crucial observation is that, despite the presence of candidate samples exhibiting similar features to the ground truth in both image and caption aspects, \algname can accurately identify the ground truth as the most influential training data, highlighting its efficacy in pinpointing the key training influences in the generation of synthesized images.

Applying other influence approximators, such as \texttt{IF} and \texttt{TracIn}, to large-scale diffusion models poses a challenge in terms of efficiency, as delineated in ~\citep{wang2023evaluating}. Specifically, the \texttt{IF} needs to approximate the inverse Hessian matrix as well as calculate gradients with respect to training and test samples. However, this approach becomes problematic when it comes to the large-scale stable diffusion model (e.g., around 860 million parameters) and a large amount of training data used (e.g., around 2 billion samples) ~\citep{wang2023evaluating}. On the other hand, \texttt{TracIn} requires computing the dot product between gradients with respect to every training and test sample, leading to the scalability problem for large-scale models. Therefore, as also mentioned in a previous study ~\citep{akyurek2022tracing}, for the large-scale language model, one selects the first layer, and for the classification model, the last layer has been widely selected. \emph{However, both baselines have not been widely explored and applied to the diffusion models to achieve high efficiency and performance}.
Therefore, addressing these scalability challenges within such baselines will be deferred to future research.

\label{app:diffusion}
\begin{figure*}[ht!]
\centering
\includegraphics[width=\linewidth]{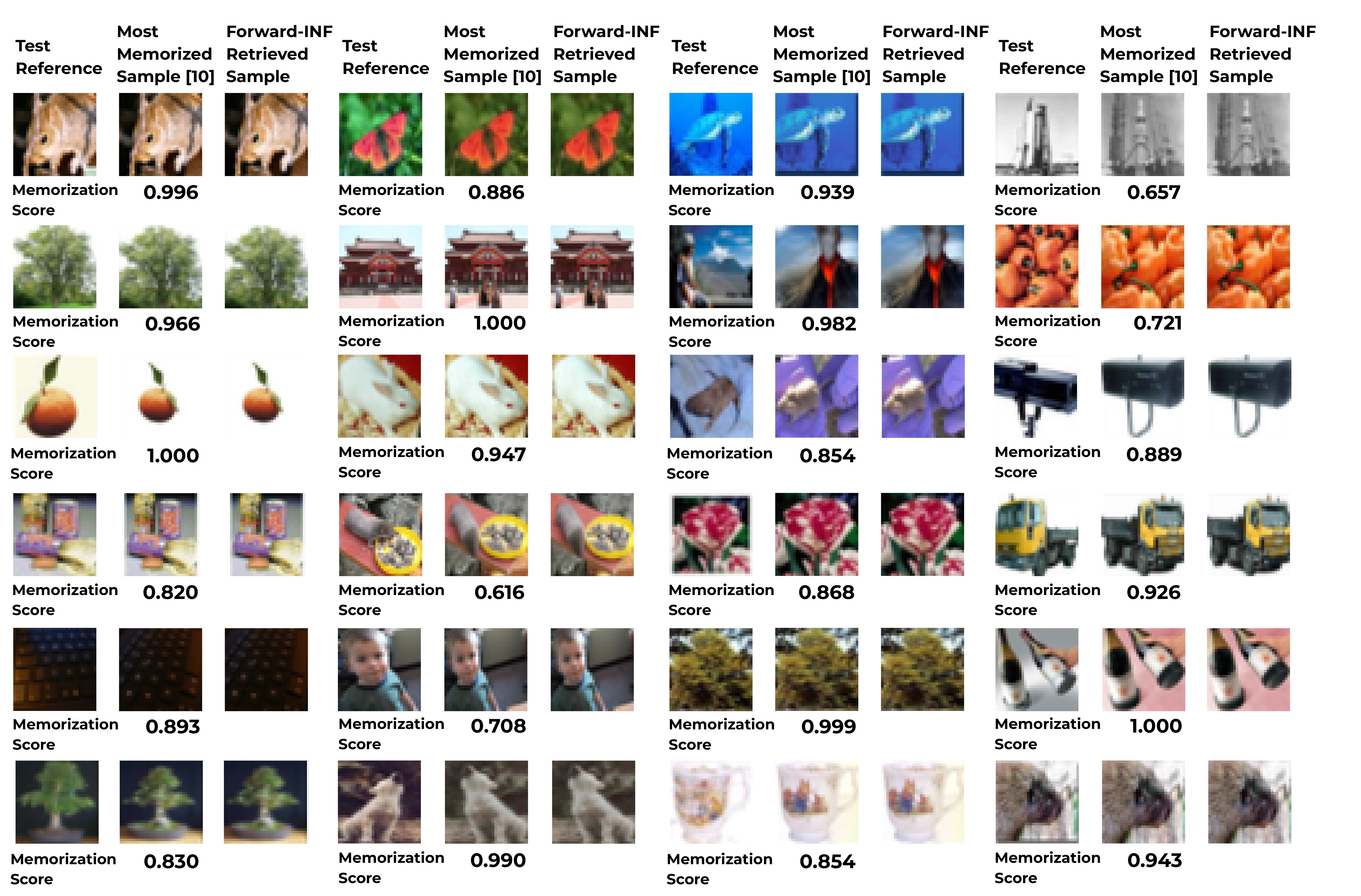}
\caption{Extensive results on memorization experiment. Similar to Figure~\ref{fig:memorization}, we present the most influential samples provided by ~\citep{feldman2020neural}, and samples that were retrieved by \algname.}
\label{fig:memorization-extensive}
\end{figure*}

\begin{table}[ht]
\centering
\renewcommand{\arraystretch}{1.5} 
\resizebox{0.9\linewidth}{!}{ 
\begin{tabular}{c|cccc|}
\textbf{Candidate Set Size}  & \textbf{100} & \textbf{1K} & \textbf{10K} & \textbf{20K} \\ 
\cmidrule[0.01pt](l{4.9em}r{4.9em}){1-1}
\textbf{T-1 Detection}  & 1.000 & 1.000 & 1.000 & 1.000 \\
\end{tabular}
}
\caption{Top-1 detection accuracy for identifying the ground truth influential sample within candidate sets of varying sizes for the stable diffusion model. \algname shows 100\% detection performance across different candidate set sizes.}
\label{tab:detection_accuracy}
\end{table}

\begin{figure*}[t!]
\centering
\includegraphics[width=.85\linewidth]{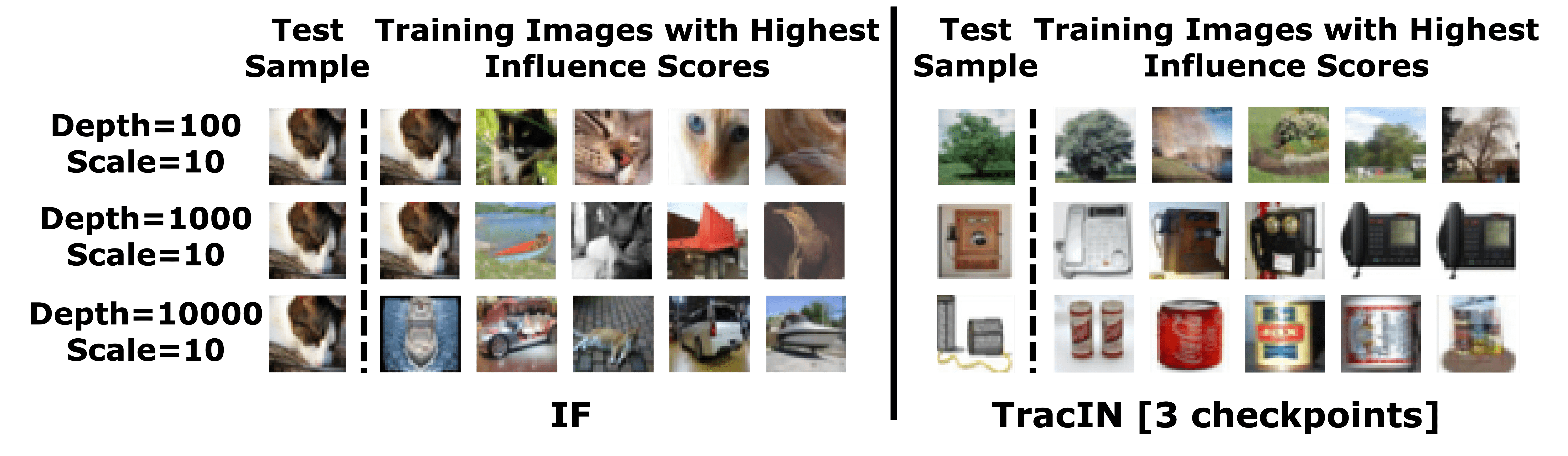}
\caption{\texttt{TracIn} and \texttt{IF} results on data leakage experiments. The right side of the figure illustrates that even though TracIn can often identify visually similar samples to the test example, it fails to detect the ground-truth leaked sample. The left side of the figure showcases the results using different depths and scales to empirically demonstrate why IF-10000 falls short.}
\vspace{-1em}
\label{fig:IF_TracIn}
\end{figure*}

\subsection{Data Leakage Detection [Section ~\ref{sec: training data leakage subsection}]}
\label{app:training data leakage subsection}
We further extend our experiment of data leakage detection on ImageNet-100 to validate the practicality as well as scalability. We observe a high detection rate (e.g., 88\% top-1 detection rate) for the RN18 classifier trained on the ImageNet-100. As the model complexity increases, the detection rate slightly decreases, considering the previous results on CIFAR-10 and CIFAR-100. We hypothesize that the unlearning process affects more diverse neurons and their connections when larger and more complex datasets are used, leading to a drop in performance. However, this performance is still considered high, compared with our baseline such as \texttt{IF} which completely fails as shown in Table {~\ref{tab:train-test more leakage}}. 

\paragraph{Unlearning dynamics.}
We additionally conduct a feature embedding analysis to visualize the gradient ascent dynamics in our method. In Figure ~\ref{fig:unlearning dynamics}, we present the feature space change according to the maximization steps. 
Interestingly, we observe that the duplicated sample (i.e., top-1) is strongly impacted by the gradient ascent process, experiencing a deviation from the class cluster during the process of unlearning. The figure in the bottom row shows examples that receive top-5 scores out of all training samples, indicating that our approach correctly identifies both the exact duplicated sample as well as relevant near duplicates. This observation supports our intuition that if we unlearn one specific point, we could expect that class or sample-related points would receive a high training loss change, thereby receiving a high \algname score.

\paragraph{Sensitivity analysis of hyperparameter selection.}
Considering that the unlearning process may involve sensitivity to hyperparameters (e.g., learning rate, and the number of iterations), our method initially selects the hyperparameter based on a holdout validation set. Consequently, we also present the results concerning the sensitivity of the validation set size. As illustrated in Table {~\ref{tab:validation size}}, \algname still achieves 100\% detection accuracy for our approach even with only 25 validation samples, indicating that our approach exhibits low sensitivity to changes in the size of the validation set.

\begin{table*}[ht]
\centering
\renewcommand{\arraystretch}{2.1}
\resizebox{0.85\linewidth}{!}{
\begin{tabular}{ccccccc}
\textbf{Model} & \textbf{Dataset} &
  \textbf{Metric} &
  \textbf{5\% {[}25 samples{]}} &
  \textbf{10\% {[}50 samples{]}} &
  \textbf{20\% {[}100 samples{]}} &
  \textbf{30\% {[}150 samples{]}} \\ 
  \cmidrule[0.4pt](l{0.525em}r{0.525em}){1-1}
 \cmidrule[0.4pt](l{0.525em}r{0.525em}){2-2} \cmidrule[0.4pt](l{0.525em}r{0.525em}){3-3} \cmidrule[0.4pt](l{0.525em}r{0.525em}){4-4} \cmidrule[0.4pt](l{0.525em}r{0.525em}){5-5} \cmidrule[0.4pt](l{0.525em}r{0.525em}){6-6} \cmidrule[0.4pt](l{0.525em}r{0.525em}){7-7}
ResNet-18 & CIFAR-100 &
  T-1 &
  1.000 &
  1.000 &
  1.000 &
  1.000 \\
ResNet-18 & CIFAR-100 &
  T-5 &
  1.000 &
  1.000 &
  1.000 &
  1.000 \\ 
  \cmidrule[0.4pt](l{0.525em}r{0.525em}){1-1}
 \cmidrule[0.4pt](l{0.525em}r{0.525em}){2-2} \cmidrule[0.4pt](l{0.525em}r{0.525em}){3-3} \cmidrule[0.4pt](l{0.525em}r{0.525em}){4-4} \cmidrule[0.4pt](l{0.525em}r{0.525em}){5-5} \cmidrule[0.4pt](l{0.525em}r{0.525em}){6-6} \cmidrule[0.4pt](l{0.525em}r{0.525em}){7-7}
\end{tabular}
}
\caption{Data leakage detection performance using ResNet-18 across various validation set sizes for the sensitivity analysis of hyperparameter selection. \algname consistently delivers high performance, even with small validation set sizes.}
\label{tab:validation size}
\end{table*}

\subsection{Memorization Analysis [Section ~\ref{sec: memorization}]}
\label{app:memorization}

We provide extensive qualitative results for memorization analysis in Figure~\ref{fig:memorization-extensive}. As we can observe from the figure, our \algname can effectively retrieve the most influential point with a much lower cost than the previous work~\citep{feldman2020neural}. In addition, based on qualitative results, we consistently observe that the high-influential pairs are exact or near duplicated samples and benefit the most from memorization (see the corresponding memorization score from the figure). 


%

\subsection{Model Behavior Tracing [Section ~\ref{sec: model behavior tracing}]}
\label{app:tracing model behavior}

In this section, we delve deeper into the problem of model behavior tracing, expanding the scope of our discussion. In reality, it becomes inherently challenging to locate direct supporting samples from the training data that precisely correspond to a given test sample. Instead, the relationships between training samples and generating answers for the test example often involve more indirect correlations. In light of this, we introduce an additional experimental design to explore a scenario wherein the test query undergoes paraphrasing. The key question is whether we can still identify the relevant ground-truth samples from the training data.

As demonstrated in Table~\ref{tab: paraphrase behavior tracing}, our proposed approach, \algname, consistently displays favorable performance, albeit with a minor drop, while outperforming the baseline \texttt{TracIn} in terms of MRR and Precision metrics for the 15K candidate set size.

\begin{table*}[ht]
\centering
\renewcommand{\arraystretch}{1.9}
\resizebox{0.65\linewidth}{!}{
\begin{tabular}{lccccc}
  \multirow{2}{*}{\begin{tabular}[c]{@{}l@{}}\textbf{Candidate Set Size}\\ \textbf{Metric}\end{tabular}}    & \multicolumn{4}{c}{\textbf{15k}}                 & \textbf{Inspected Queries} \\ 
\cmidrule[0.4pt](l{0.525em}r{0.525em}){2-5} \cmidrule[0.4pt](l{0.525em}r{0.525em}){6-6}
 & \textbf{MRR} & \textbf{$\Delta$} & \textbf{Precision} & \textbf{$\Delta$} & \textbf{\# of Queries/Min}          \\ 
\cmidrule[0.4pt](l{0.525em}r{0.525em}){1-1}
 \cmidrule[0.4pt](l{0.525em}r{0.525em}){2-2} \cmidrule[0.4pt](l{0.525em}r{0.525em}){3-3} \cmidrule[0.4pt](l{0.525em}r{0.525em}){4-4} \cmidrule[0.4pt](l{0.525em}r{0.525em}){5-5} \cmidrule[0.4pt](l{0.525em}r{0.525em}){6-6}
\texttt{TracIn} (single)         & 0.1868 & -      & 0.1549 & -      & 396.125           \\
\textbf{\algname} & \textbf{0.2031} & \textbf{0.0163} & \textbf{0.1551} & \textbf{0.0002} & \textbf{1289.475} \\ 
\cmidrule[0.4pt](l{0.525em}r{0.525em}){1-1}
 \cmidrule[0.4pt](l{0.525em}r{0.525em}){2-2} \cmidrule[0.4pt](l{0.525em}r{0.525em}){3-3} \cmidrule[0.4pt](l{0.525em}r{0.525em}){4-4} \cmidrule[0.4pt](l{0.525em}r{0.525em}){5-5} \cmidrule[0.4pt](l{0.525em}r{0.525em}){6-6}
\end{tabular}
}
\vspace{0.5em}
\caption{Behavior tracing performance comparison of different attribution methods on the paraphrased candidate set.}
\label{tab: paraphrase behavior tracing}
\end{table*}

\begin{figure*}[ht]
\centering
\includegraphics[width=0.9\linewidth]{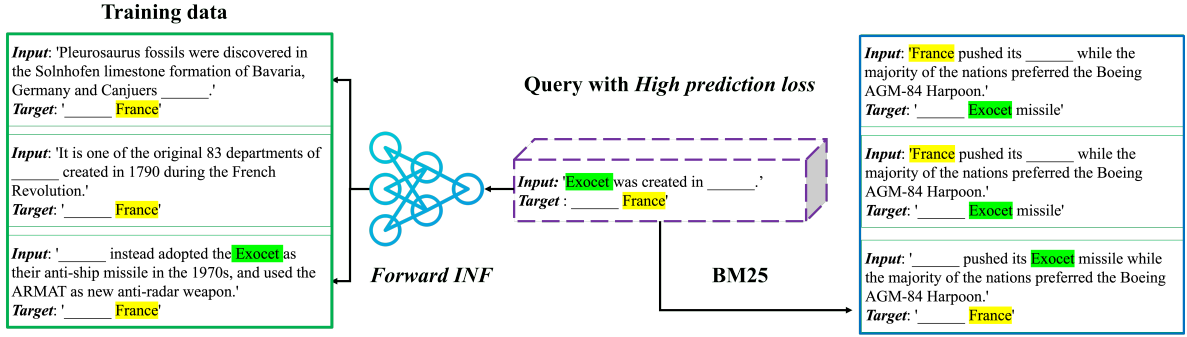}
\caption{Comparison of the top-3 selected training samples from \algname and BM25.}
\vspace{-1.5em}
\label{fig:behavior tracing}
\end{figure*}

\noindent\paragraph{Comparison with simple model-independent information retrieval baseline~\citep{robertson1995okapi}.} BM25 is a standard information retrieval technique that selects proponents by retrieving training examples with high lexical overlap with the query. BM25 has been adopted to trace facts in~\citep{akyurek2022tracing}.  However, since the influence scores derived from BM25 do not incorporate information pertinent to the specific model under consideration, it \emph{cannot} be used to explain why a \emph{given model} makes some factual assertions. 
For instance, when our LLM produces an incorrect prediction resulting in a high loss, we want to trace back to the contributing examples that lead to this misjudgment for the sake of transparency and interpretability. While the performance of BM25 in identifying pertinent facts is promising ~\citep{akyurek2022tracing}, it falls short in shedding light on the intricate model's behavior (i.e., it only provides the related facts but does not explain why a model outputs high loss). Conversely, our proposed method is adept at elucidating the underlying reasons behind LLM's inaccurate answers. 
For example, as shown in Figure ~\ref{fig:behavior tracing}, the top-1 pair retrieved by our method includes the target label ``France.'' However, when considering the overall meaning of the sentence, it indicates a different semantical meaning. 
This observation implies the presence of semantic conflicts (i.e., some samples share the same labels but entail different semantical meanings) within the training samples, consequently confusing the model's learning process.


\end{document}